\documentclass[accepted]{uai2022hack} 

\usepackage[british]{babel}

\usepackage{natbib} 
\usepackage{mathtools} 
\usepackage{booktabs} 
\usepackage{tikz} 
\usepackage{amsmath, amsfonts, mathtools}

\usepackage{algorithm}
\usepackage[noend]{algpseudocode}
\usepackage{graphicx}
\usepackage{mwe}
\usepackage{graphbox} 
\usepackage{multirow}

\def\x{{\mathbf{x}}}
\def\p{{\mathbf{p}}}

\def\z{{\mathbf{z}}}

\def\y{{\mathbf{y}}}

\def\bpi{{\boldsymbol{\pi}}}

\def\bmu{{\boldsymbol{\mu}}}

\def\bE{{\mathbb{E}}}
\def\bR{{\mathbb{R}}}

\newcommand{\cut}[1]{}

\title{Inference for Generative Capsule Models}

\author[1]{Alfredo Nazabal\thanks{Equal contribution.}\thanks{Work carried out while AN was at the Alan Turing Institute.}}
\author[2]{Nikolaos Tsagkas$^\ast$ \thanks{Part of this work was carried out
    when NT was a MSc student at the University of Edinburgh.}}
\author[3,4]{Christopher K. I. Williams}
\affil[1]{%
    Amazon Development Centre Scotland,
    Edinburgh, UK
}
\affil[2]{%
  \texttt{tsagkas.nikolas@gmail.com}
}
\affil[3]{%
  School of Informatics,
  University of Edinburgh,
  Edinburgh EH8 9AB, UK
}
\affil[4]{%
  The Alan Turing Institute,
  London, UK
}  

\begin{document}
\maketitle

\begin{abstract}
Capsule networks (see e.g.~\citealp{hinton2018matrix}) aim to encode
knowledge and reason about the relationship between an object and its
parts. In this paper we specify a \emph{generative} model for such data, and
derive a variational algorithm for inferring the transformation of
each object and the assignments of observed parts to the objects. We
apply this model to (i) data generated from multiple geometric objects
like squares and triangles (``constellations''), and (ii) data from a
parts-based model of faces.
Recent work by~\cite{kosiorek2019stacked} has used
amortized inference via stacked capsule autoencoders (SCAEs) to tackle
this problem---our results show that we significantly outperform them where
we can make comparisons (on the constellations data).
\end{abstract}

\section{Introduction}
An attractive way to set up the problem of object recognition is
\emph{hierarchically}, where an object is described in terms of its
parts, and these parts are in turn composed of sub-parts, and so on.
For example a face can be described in terms of the eyes, nose, mouth,
hair, etc.; and a teapot can be described in terms of a body, handle,
spout and lid parts.  This approach has a long history in computer
vision, see e.g.\ Recognition-by-Components by \citet{biederman-87}.
Advantages of recognizing objects by first recognizing their constituent
parts include tolerance to the occlusion of some parts, and that parts
may vary less under a change of pose than the appearance of the whole
object.

One big advantage of a parts-based setup is that if the
pose\footnote{i.e.\ the location and rotation of the object in 2D or
3D.} of the object changes, this can have very complicated effects on
the pixel intensities in an image, but the geometric transformation
between the object and the parts can be described by a simple linear
transformation (as used in computer graphics).  Convolutional neural
networks can allow a recognition system to handle 2D shifts of an object
in the image plane, but attempts to generalize such equivariances
across more general transformations lead to cumbersome constructions,
and have not been very successful.

Recent work by Hinton and
collaborators~\citep{sabour2017dynamic,hinton2018matrix} has developed
\emph{capsule networks}.  The key idea here is that a part in a lower
level can vote for the pose of an object in the higher level, and an
object's presence is established by the agreement between votes for
its pose.  \citet[p.~1]{hinton2018matrix} use an iterative process
called ``routing-by-agreement'' which ``updates the probability with
which a part is assigned to a whole based on the proximity of the vote
coming from that part to the votes coming from other parts that are
assigned to that whole''.  Subsequently~\citet{kosiorek2019stacked}
framed inference for a capsule network in terms of an autoencoder, the
Stacked Capsule Autoencoder (SCAE).  Here, instead of the iterative
routing-by-agreement algorithm, a neural network $h^{\mathrm{caps}}$
takes as input the set of input parts and outputs predictions for the
object capsules' instantiation parameters $\{\y_k \}_{k=1}^K$.
Further networks $h_k^{\mathrm{part}}$ are then used to predict part
candidates from each $\y_k$.

The objective function used in~\citet{hinton2018matrix} (their
eq.\ 4) is quite complex (involving four separate terms), and is not
derived from first principles.
In this paper we argue that the description in the paragraph above is backwards---it is more natural to describe the generative
process \emph{by which an object gives rise to its parts}, and that the appropriate routing-by-agreement inference algorithm then falls out naturally from this principled formulation.
Below we focus on a single-layer of part-object relations, as we need
to show first that this is working properly; but the generative model
should be readily extensible to deeper hierarchies.

The contributions of this paper are to:
\begin{itemize}
  \item  Derive a variational inference algorithm for
    a generative model of object-part relationships,
    including a relaxation of the permutation-matrix formulation for matching
    object parts to observations;
  \item Obtain a principled and novel routing-by-agreement algorithm from this formulation;
  \item Demonstrate the effectiveness of our method on (i)
    ``constellations'' data generated from multiple geometric objects
    (e.g. triangles, squares) at arbitrary translations, rotations and
    scales;  and (ii) data from a novel parts-based model of faces.
  \item Evaluate the performance of our model vs.\ competitors on the
    constellations data.
\end{itemize}

\begin{figure}[t]
    \centering
    \includegraphics[width=0.7\linewidth]{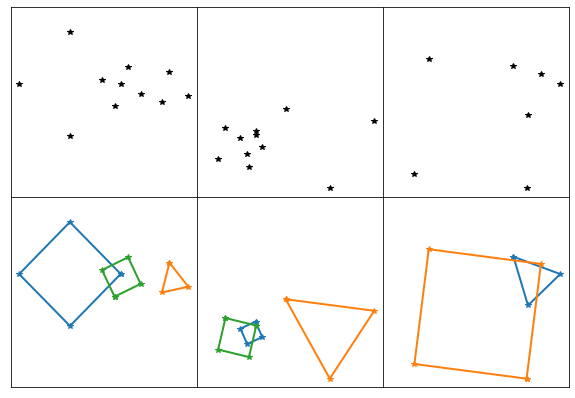}
\caption{Scenes composed of 2D points (upper
figures) and their corresponding objects (lower
figures).}
    \label{fig:constellations}
\end{figure}

\section{Overview}
Our images consist of a set of objects in different poses. Examples
include images of handwritten digits, faces, or geometric shapes in 2D
or 3D.  An object can be defined as an instantiation of a specific
model object (or template) along with a particular pose (or geometric
transformation).  Furthermore, objects, and thus templates, are
decomposed in parts, which are the basic elements that comprise the
objects.  For example,
faces can be decomposed into parts (e.g.\ mouth, nose etc.),
or a geometric shape can be decomposed into vertices. These parts
can have internal variability, (e.g.\ eyes open or shut).

More formally, let $T = \{T_k\}_{k=1}^K$ be the set of $K$ templates
that are used to generate a scene.  Each template
$T_k=\{\p_n\}_{n=1}^{N_k}$ is composed of $N_k$ parts $\p_n$.  We
assume that scenes can only be generated using the available
templates. Furthermore, every scene can present a different
configuration of objects, with some objects missing from some scenes.
For example, in scenes that could potentially contain all digits from
0 to 9 once, and if only the digits 2 and 5 are in the image, we consider
that the other digits are missing. If all the templates were
employed in the scene, then the number of observed parts $M$ is equal
to the sum of all the parts of all the templates $N = \sum_{k=1}^K
N_k$.

Each observed template $T_k$ in a scene is then transformed by an
independent transformation $\y_k$, different for each template,
generating a transformed object $X_k = \{\x_n\}_{n=1}^{N_k}$
\begin{equation}
T_k \xrightarrow{\y_k} X_k .
\end{equation}
The transformation $\y_k$ includes a geometric transformation of
the template, but also includes appearance variability in the parts.

Assume that we are given a scene $X = \{\x_m\}_{m=1}^M$ composed of
$M$ observed parts coming from multiple templates.
The \emph{inference problem} for $X$ involves a number of 
different tasks.  We need to determine which objects from the set of
templates were employed to generate the scene. 
Also, we need to infer what transformation
$\y_k$ was applied to each template to generate the objects.  This
allows us to infer the \emph{correspondences} between the
template parts and the scene parts.

We demonstrate our method on ``constellations'' data as shown in
Fig.\ \ref{fig:constellations}, and data from a parts-based model of faces
illustrated in Fig.\ \ref{fig:face_example}.

\section{Generative model for capsules}\label{sec:generative}
We propose a generative model to describe the problem.
Consider a template (or model) $T_k$ for the $k$th object. $T_k$ is
composed of $N_k$ parts $\{ \p_n \}_{n=1}^{N_k}$.\footnote{For
simplicity of notation we suppress the dependence of $\p_n$ on $k$ for now.}
Each part $\p_n$ is described in its reference frame by its \emph{geometry} $\p^g_n$
and its \emph{appearance} $\p^a_n$.  Each object also has associated
latent variables $\y_k$ which transform from the reference frame to the
image frame, so $\y_k$ is split into geometric variables $\y^g_k$ and
appearance variables $\y^a_k$.

\paragraph{Geometric transformations:}
Here we consider 2D templates and a similarity transformation
(translation, rotation and scaling) for each object, but this can be
readily extended to allow 3D templates and a scene-to-viewer camera
transformation. We assume that $\p_n^g$ contains the $x$ and $y$
locations of the part, and also its size $s_n$ and orientation
$\phi_n$ relative to the reference frame.\footnote{For
the constellations data, the size and orientation information is not
present, nor are there any appearance features.}
The size and orientation are represented as the projected size of the
part onto the $x$ any $y$ axes, as this allows us to use linear
algebra to express the transformations (see below).
Thus $\p_n^g = (p^g_{nx}, p^g_{ny}, s_n \cos\phi_n, s_n \sin\phi_n)^T$.

Consider a template with parts $\p^g_n$ for $n = 1,
\ldots, N_k$ that we wish to scale by a factor $s$, rotate through
with a clockwise rotation angle $\theta$ and translate by $(t_x,t_y)$.
We obtain a transformed object with geometric observations for the
$n$th part $\x^g_n = (x^g_{nx},x^g_{ny},x^g_{nc}, x^g_{ns})$, where the $c$ and $s$
subscripts denote the projections of the scaled and rotated part onto
the $x$ and $y$ axes respectively ($c$ and $s$ are mnemonic for cosine and sine).

For each part in the template, the geometric transformation works as follows:
\begin{equation}\label{eq:nxycs}
\begin{pmatrix} x_{nx}\\ x_{ny} \\ x_{nc} \\ x_{ns}\end{pmatrix} =   
\begin{pmatrix} 1 & 0 & p_{nx} & p_{ny}\\
         0 & 1 & p_{ny} & -p_{nx} \\
         0 & 0 & s_n \cos\phi_n & - s_n \sin\phi_n \\
         0 & 0 & s_n \sin\phi_n & s_n \cos\phi_n \\    \end{pmatrix}
\begin{pmatrix}
t_x\\
t_y\\
s\cos{\theta}\\
s\sin{\theta}
\end{pmatrix}.
\end{equation}
Decoding the third equation, we see that $x_{nc} = s_n s
\cos\phi_n \cos \theta - s_n s \sin\phi_n \sin \theta =
s_n s \cos(\phi_n + \theta)$ using standard trigonometric identities.
The $x_{ns}$ equation is derived similarly. We shorten eq.\ \ref{eq:nxycs} to
$\x^g_n = F^g_{kn} \y^g_k$,
where $\y^g_k$ is the $\bR^4$ column vector, and $F^g_{kn} \in \bR^{4 \times 4}$
is the matrix to its left. Allowing Gaussian observation noise with
precision $\lambda$ we obtain
\begin{equation}
p(\x^g_n|T_k, \y^g_k) \sim
N\left(\x^g_{n}|F^g_{kn}\y^g_{k},\lambda^{-1}I\right).
\label{eq:pxg_supp}
\end{equation}

The prior distribution over similarity transformations $\y^g_k$ is
modelled with a $\bR^4$ Gaussian distribution with mean $\bmu^g_0$ and
covariance matrix $D^g_0$: 
\begin{equation}
\label{eq:p_Y}
p(Y^g) = \prod_{k=1}^K N(\y^g_{k}|\bmu^g_0,D^g_0).
\end{equation}
Notice that modelling $\y^g_k$ with a Gaussian distribution implies that
we are modelling the translation $(t_x,t_y)$ in $\bR^2$ with a
Gaussian distribution.  If $\bmu^g_0=\mathbf{0}$ and $D^g_0 = I_4$ 
then $s^2 = (y^g_{k3})^2 + (y^g_{k4})^2$
has a $\chi^2_2$ distribution, and
$\theta = \arctan{{y^g_{k4}}/{y^g_{k3}}}$ is uniformly distributed
in its range $[-\pi,\pi]$ by
symmetry.  For more complex linear transformations (e.g.\ an affine
transformation), we need only to increase the dimension of $\y^g_k$ and
change the form of $F^g_{kn}$, but the generative model
in~\eqref{eq:p_Y} would remain the same.

\paragraph{Appearance transformations:}
The appearance $\x_n^a$ of part $n$ in the image depends on 
$\y^a_k$. For our faces data, $\y^a_k$ is a vector latent variable
which models the co-variation of the appearance of the parts via
a linear (factor analysis) model; see sec.\ \ref{sec:parts_face} for a fuller
description.
Hence
\begin{equation}
p(\x^a_n|T_k, \y_k) \sim
N\left(\x^a_{n}|F^a_{kn} \y^a_{k} + \boldsymbol{m}_{kn}^a, D^a_{kn}\right),
\label{eq:pxa_supp} 
\end{equation}
where $F^a_{kn}$ maps from $\y^a_k$ to the predicted appearance features
in the image, $D^a_{kn}$ is a diagonal matrix of variances and
$\boldsymbol{m}_{kn}^a$ allows for the appearance features to have a
non-zero mean. The dimensionality of the $n$th part of the $k$th
template is $d_{kn}$. The prior for $\y^a_k$ is taken to be a standard
Gaussian, i.e.\ $N(\mathbf{0},I)$. Combining (\ref{eq:p_Y}) and the
prior for $\y^a_k$, we have that $p(\y_k)  = N(\bmu_0, D_0)$, where
$\bmu_0$ stacks $\bmu^g_0$ and $\mathbf{0}$ from the appearance,
and $D_0$ is a diagonal matrix with blocks $D^g_0$ and $I$.

\paragraph{Joint distribution:}
Let $z_{mnk} \in \{0,1\}$ indicate whether observed part $\x_m$
matches to part $n$ of object $k$.
Every observation $m$ belongs uniquely to a tuple $(k,n)$, or in other
words, a point $\x_m$ belongs uniquely to the part defined by $\y_k$
acting on the template matrix $F_{kn}$. The opposite is also partially true;
every tuple $(k,n)$ belongs uniquely to a point $m$ or it is
unassigned if part $n$ of template $k$ is missing in the scene.

The joint distribution of the variables in the model is given by
\begin{equation}
p(X,Y,Z) = p(X|Y,Z)p(Y)p(Z),
\end{equation}
where $p(X|Y,Z)$ is a Gaussian mixture model explaining how the points in a scene were generated from the templates
\begin{equation}\label{eq:p_X}
p(X|Y,Z) = \prod_{m=1}^M\prod_{k=1}^K\prod_{n=1}^{N_k}
N\left(\x_{m}|F_{kn}\y_{k} + \boldsymbol{m}_{kn}, D_{kn} \right)^{z_{mnk}},
\end{equation}
where $D_{kn}$ consists of the diagonal matrices
$\lambda^{-1} I$ and $D^a_{kn}$ and $\boldsymbol{m}_{kn}$
consists of a zero vector for the geometric features stacked
on top of the mean for the appearance features
$\boldsymbol{m}^a_{kn}$.
Note that $F_{kn}$ has blocks of
zeros so that $\x^g_m$ does not depend on $\y^a_{k}$,
and $\x^a_m$ does not depend on $\y^g_{k}$.
Note that such a model for a \emph{single object} with geometric
features only was developed
by~\citet{revow-williams-hinton-96}.

\paragraph{Annealing parameter:} During the fitting of the model to data, it is useful to modify the covariance matrix $D_{kn}$ to $\beta^{-1} D_{kn}$, where $\beta$ is a parameter $< 1$. The effect of this is to inflate the variances in $D_{kn}$, allowing greater uncertainty in the inferred matches early on in the fitting process, as used, e.g.\ in \citet{revow-williams-hinton-96}.  $\beta$ is increased according to an annealing schedule during the fitting.

\paragraph{Match distribution $p(Z)$:}
In a standard Gaussian mixture model, the assignment matrix $Z$ is
characterized by a Categorical distribution, where each point $\x_m$
is assigned to one part
\begin{equation}
\label{eq:p_Z}
p(Z) = \prod_{m=1}^M \text{Cat}(\z_{m}|\bpi),
\end{equation}
with $\z_m$ {being a 0/1 vector with only one 1,}
and $\bpi$ being the probability vector for each tuple $(k,n)$.
However, the optimal solution to our problem occurs when each part of
a template belongs uniquely to one observed part in a scene.  This means that
$Z$ should be a permutation matrix, where each point $m$ is assigned
to a tuple $(k,n)$ and vice versa.
Notice that a permutation matrix is a square matrix, so if $M \leq N$,
we add dummy rows to $Z$, which are assigned to missing points in the
scene.

The set of permutation matrices of dimension $N$ is a discrete set
  containing $N!$ permutation matrices. They can be modelled with a
  discrete prior over permutation matrices, assigning each matrix
  $Z_i$ a probability $\pi_i$, such as:
\begin{equation}
\label{eq:p_Z_true}
p_{perm}(Z) = \sum_{i=1}^{N!}  \pi_i \; I[Z=Z_i] 
\end{equation}
with $\sum_{i=1}^{N!} \pi_i = 1$ and $I[Z=Z_i]$ the indicator
function, being equal to $1$ if $Z=Z_i$ and $0$ otherwise.

The number of possible permutation matrices increases
as $N!$, {which makes exact inference over permutations intractable.}
An interesting property of $p_{perm}(Z)$ is that its first moment
$\bE_{p_{perm}}[Z]$ is a doubly-stochastic (DS) matrix, a matrix of
elements in $[0,1]$ whose rows and columns sum to 1.
We propose to relax $p_{perm}(Z)$ to a distribution $p_{DS}(Z)$ that
is characterized by the doubly-stochastic matrix $A$ with elements
$a_{mnk}$, such that $\bE_{p_{DS}}[Z]=A$:
\begin{equation}
\label{eq:p_Z_approx}
p_{DS}(Z) =  \prod_{m=1}^{N}\prod_{k=1}^{K}\prod_{n=1}^{N_k} a_{mnk}^{z_{mnk}}.
\end{equation}
$A$ is fully characterized by $(N-1)^2$ elements. In the absence of
any prior knowledge of the affinities, 
a uniform prior over $Z$ with $a_{mnk}=\frac{1}{N}$ can be used.
However, note that $p_{DS}$ can
also represent a particular permutation matrix $Z_i$ by setting the
appropriate entries of $A$ to 0 or  1, and indeed we would expect this to occur during
variational inference (see sec.\ \ref{sec:VI}) when the model
converges to a correct solution.

\section{Variational Inference}\label{sec:VI}
Variational inference for the above model can be derived similarly to the Gaussian Mixture model case~\citep[Chapter 10]{bishop2006pattern}.
The variational distribution under the mean field assumption is given by
$q(Z,Y) = q(Z)q(Y)$,
%
where the optimal solutions for $\log{q(Z)}$ and $\log{q(Y)}$ under the generative model can be expressed as
\begin{eqnarray}
\log{q(Z)} &\propto & \bE_{q(Y)}[\log{p(X,Y,Z)}] ,\\
\log{q(Y)} &\propto & \bE_{q(Z)}[\log{p(X,Y,Z)}] .
\end{eqnarray}
For $q(Z)$ we obtain an expression with the same distribution model as the prior in~\eqref{eq:p_Z_approx}
\begin{eqnarray}
q(Z) \propto \prod_{m=1}^N\prod_{k=1}^K\prod_{n=1}^{N_k} \rho_{mnk}^{z_{mnk}},\label{eq:q_Z}
\end{eqnarray}
where $\rho_{mnk}$ represents the unnormalized probability of point $m$ being assigned to tuple $(k,n)$ and vice versa.
These unnormalized probabilities have a different form depending on whether we are considering a point that appears in the scene ($m \leq M$)
\begin{align}
&\log{\rho_{mnk}} = \log{a_{mnk}}-\frac{1}{2}\log{|\beta^{-1} D_{kn}|}- \frac{d_{kn}}{2}\log{2\pi} - \nonumber\\
&\frac{\beta}{2}\bE_{\y_{k}}[(\x_{m}-F_{kn}\y_{k}-\boldsymbol{m}_{kn})^T D_{kn}^{-1} (\x_{m}-F_{kn}\y_{k}-\boldsymbol{m}_{kn})],\label{eq:log_rho}
\end{align}
or whether we are considering a dummy row of the prior ($m > M$),
\begin{align}
\log{\rho_{mnk}} &= \log{a_{mnk}},\label{eq:log_rho_dummy}
\end{align}
When a point is part of the scene~\eqref{eq:log_rho}, and thus $m \leq M$ the update of $\rho_{mnk}$ is similar to the Gaussian mixture model case.
However, if a point is not part of the scene~\eqref{eq:log_rho_dummy}, and thus $m > M$ then the matrix is
not updated and the returned value is the prior $a_{mnk}$.
This specific choice of distribution for both $p_{DS}(Z)$ and $q(Z)$ leads
naturally to the addition of dummy rows that are employed to make
the matrix $\rho$ square.
{Now, the normalized distribution $q(Z)$ becomes:
\begin{eqnarray}
q(Z) = \prod_{m=1}^N\prod_{k=1}^K\prod_{n=1}^{N_k} r_{mnk}^{z_{mnk}},\label{eq:R_doubleStochastic}
\end{eqnarray}
where $\bE_{q(Z)}[z_{mnk}] = r_{mnk}$.
The elements $r_{mnk}$ of matrix $R$ represent the posterior probability of each point $m$ being uniquely assigned to the part-object tuple $(n,k)$ and vice-versa.
This means that $R$ needs to be a DS matrix.
{This can be achieved by employing the
  Sinkhorn-Knopp algorithm~\citep{sinkhorn1967concerning}, which
updates a square non-negative matrix by normalizing the rows and columns
alternately, until the resulting matrix becomes doubly stochastic
(see supp.\ mat.\ \ref{sec:appVI} for more details).
The use of the Sinkhorn-Knopp algorithm for approximating matching
problems has also been described by \cite{powell2019computing} and 
\cite{mena2020sinkhorn}, but note that in our case we also need to alternate
with inference for $q(Y)$.}
Furthermore, the optimal solution to the assignment problem occurs
when $R$ is a permutation matrix itself. When this happens we
exactly recover a discrete posterior (with the same form
as~\eqref{eq:p_Z_true}) over permutation matrices where one of them
has probability one, with the others being zero.

The distribution for $q(Y)$ is a Gaussian with
\begin{align}
q(Y) &= \prod_{k=1}^K N(\y_k|\bmu_k,\Lambda_k^{-1}), \\\label{eq:Lambda_k}
\Lambda_{k} &= D^{-1}_{0} +\beta \sum^M_m \sum^{N_k}_n r_{mnk}F_{kn}^{T}D_{kn}^{-1} F_{kn}, \\\label{eq:mu_k}
\bmu_{k} &= \Lambda_k^{-1} \left[D_{0}\bmu_{0} + \beta\sum^M_m\sum^{N_k}_nr_{mnk}F_{kn}^{T}D_{kn}^{-1}(\x_{m}-\boldsymbol{m}_{kn})\right],
\end{align}
where the updates for both $\Lambda_k$ and $\bmu_{k}$ depend explicitly
on the annealing parameter $\beta$ and the templates employed in the
model.
Note that the prediction from datapoint $m$ to the mean of
$\y_k$ is given by
$r_{mnk}F_{kn}^{T}(D_{kn})^{-1}(\x_{m}-\boldsymbol{m}_{kn})$, i.e.\ a
weighted sum of the predictions of each part $n$ with weights
$r_{mnk}$.
These expressions remain the same when considering a Gaussian mixture prior such as~\eqref{eq:p_Z}.

The expectation term in~\eqref{eq:log_rho} under the model and
evidence lower error bound (ELBO) are given in
supp.\ mat. \ref{sec:appVI}.
Algorithm~\ref{alg:variational_inference} summarizes the
inference procedure for this model; the alternating updates of $q(Y)$
and $q(Z)$ carry out routing-by-agreement.

\begin{algorithm}[t]
\caption{Variational Inference}
\label{alg:variational_inference}
\begin{algorithmic}[1]
\State Initialize $\beta$, $\beta_{max}$ and $R \sim U[0,1]^{N\times N}$, $\forall m,n,k$
\State $R = \text{SinkhornKnopp}(R)$
\While{not converged}
	\State Update $\Lambda_k$~\eqref{eq:Lambda_k}, $\forall k$
	\State Update $\bmu_k$~\eqref{eq:mu_k}, $\forall k$
	\State Update $\log{\rho_{mnk}}$~\eqref{eq:log_rho}\eqref{eq:log_rho_dummy}, $\forall m,n,k$
	\State Update $R = \text{SinkhornKnopp}(\rho)$
	\If{ELBO has converged}
		\If{$\beta < \beta_{max}$}
			\State Anneal $\beta$
		\Else
			\State converged = True
		\EndIf
	\EndIf
\EndWhile
\Return $R, \{ \bmu_{k},\Lambda_k \}$
\end{algorithmic}
\end{algorithm}
\subsection{Comparison with other objective functions}
In \cite{hinton2018matrix} an objective function $cost^h_k$ is defined
(their eq.\ 1) which considers inference for the pose of a
higher-level capsule $k$ on pose dimension $h$.  Translating to our
notation, $cost^h_k$ combines the votes from each datapoint $m$ as
$cost^h_k = \sum_m r_{mk} \ln P^h_{m|k}$, where $P^h_{m|k}$ is a
Gaussian, and $r_{mk}$ is the ``routing softmax assignment'' between
$m$ and $k$.  It is interesting to compare this with our equation
\eqref{eq:mu_k}.  Firstly, note that the vote of $\x_m$ to part $n$ in
object $k$ is given explicitly by $F^T_{kn} \x_m$, i.e.\ we do not
require introduction of an independent voting mechanism, this falls
out directly from the inference.  Secondly, note that our $R$ must
keep track not only of assignments to objects, but also to parts of
the objects.  In contrast to \citet{hinton2018matrix}, our inference
scheme is derived from variational inference on the generative model,
rather than introducing an \emph{ad hoc} objective function.

The specialization of the SCAE method of \citet{kosiorek2019stacked}
  to constellation data is called the ``constellation capsule
  autoencoder (CCAE)'' and discussed in their sec.\ 2.1.
Under their equation 5, we have that
\begin{align}\label{eq:CCAE_loss}
p(\x_{1:M}) = \prod_{m=1}^M\sum_{k=1}^K\sum_{n=1}^N \frac{a_ka_{k,n}}{\sum_i a_i \sum_j a_{ij}} N(\x_m|\mu_{k,n},\lambda_{k,n}),
\end{align}
where $a_k \in [0,1]$ is the presence probability of capsule $k$,
$a_{k,n} \in [0,1]$ is the conditional probability that a given
candidate part $n$ exists, and $\mu_{k,n}$ is the predicted location
of part $k,n$.
The $a_k$s are predicted by the network $h^{\mathrm{caps}}$, while the
$a_{k,n}$s and $\mu_{k,n}$s are produced by separate networks
$h_k^{\mathrm{part}}$ for each part $k$.

We note that \eqref{eq:CCAE_loss} provides an autoencoder style
reconstructive likelihood for $\x_{1:M}$, as the $a$'s and $\mu$'s
depend on the data.  To handle the arbitrary number of datapoints $M$,
the network $h^{\mathrm{caps}}$ employs a Set Transformer
architecture~\citep{lee-lee-kosiorek-choi-teh-19}.  In comparison to
our iterative variational inference, the CCAE is a ``one shot''
inference mechanism.  This may be seen as an advantage, but in scenes
with overlapping objects,
humans may perform reasoning like ``if that point is one of the
vertices of a square, then this other point needs to be explained by a
different object'' etc, and it may be rather optimistic to believe
this can be done in a simple forward pass.  Also, CCAE cannot exploit
prior knowledge of the geometry of the objects, as it relies on an
opaque network $h^{\mathrm{caps}}$ which requires extensive training.

\subsection{Related work}
Above we have discussed  the work of \citet{sabour2017dynamic,hinton2018matrix} and
\citet{kosiorek2019stacked}. There is some more recent work, for example
\citet{li-zhu-naud-xi-20}
present an undirected model, based on the capsule RBM of \citet{li-zhu-19}.
Compared to our work, this is an undirected as
opposed to directed generative model.
Also \citet{smith-schut-gal-vanderwilk-21} propose a layered directed
generative model, where the children of higher level capsules inherit
its pose (but transformed).
Both \citet{li-zhu-naud-xi-20} and
\citet{smith-schut-gal-vanderwilk-21} consider tree-structured
relationships between parents and children, similar to
\citet{hinton-ghahramani-teh-00} and \citet{storkey-williams-03}.
However, most importantly, these recent capsule models
do not properly handle the fact that the input to a capsule should be a
\emph{set} of parts; instead in their work the first layer capsules
that interact with the input image model specific location-dependent
features/templates in the image, and their
second layer capsules have interactions with the specific first layer
capsules (e.g.\ the fixed affinity $\rho^k_{ij}$ of
\citet{smith-schut-gal-vanderwilk-21} parent $i$ in layer
$k$ for child $j$).
But if we consider a second-layer capsule that is. for example, detecting
the conjunction of the 3 strokes comprising a letter ``A'', then at
different translations, scales and rotations of the A it will need to connect
to different image level stroke features, so these connection
affinities must depend on transformation variables of the parent.

\subsection{A RANSAC approach to inference}
A radical alternative to ``routing by agreement'' inference is to make
use of a ``random sample consensus'' approach (RANSAC,
\citealp*{fischler1981random}), where a minimal number of parts are
used in order to instantiate \cut{the whole} an object. The
original RANSAC fitted just one object, but Sequential RANSAC (see,
  e.g., \citealt*{torr-98,vincent-laganiere-01}) repeatedly removes
  the parts associated with a detected object and re-runs RANSAC, so
  as to detect all objects.

For the constellations problem, we can try matching
any pair of points on one of the
templates to every possible pair of $M(M-1)$
points in the scene.  The key insight is that a pair of known points
is sufficient to compute the 4-dimensional $\y_k$ vector in the case
of similarity transformations.
Using the transformation $\hat{\y}_k$, we can then \emph{predict} the
location of the remaining parts of the template, and \emph{check} if
these are actually present.  If so, this provides evidence for the
existence of $T_k$ and $\hat{\y}_k$.  After considering the $M(M-1)$
subsets, the algorithm then combines the identified instantiations to
give an overall explanation of the scene. Details of RANSAC
for the constellations problem are given in
supp.\ mat.\ \ref{sec:supp_RANSAC}.

For the faces data, each part has location, scale and orientation
information, so a single part is sufficient to instantiate the whole
object geometrically. For the appearance, we follow the procedure
for inference in a factor analysis model with missing data, as
given in \citet{williams-nash-nazabal-19}, to predict $\y^a_k$ given
the appearance of the single part.

\section{Experimental details}
Below we provide details of the data generators, inference
methods and evaluation criteria for the constellations data in sec.\
\ref{sec:expts_constellation}, and the faces data in sec.\
\ref{sec:parts_face}.

\subsection{Constellation data} \label{sec:expts_constellation}
In order to allow fair comparisons, we use the same dataset generator
for geometric shapes employed by~\cite{kosiorek2019stacked}.  We
create a dataset of scenes, where each scene consists of a set of 2D
points, generated from different geometric shapes.  The possible
geometric shapes (templates) are a square and an isosceles triangle,
with parts being represented by the 2D coordinates of the vertices.
We use the same dimensions for the templates as used
by~\cite{kosiorek2019stacked}, side 2 for the square, and base and
height 2 for the triangle.  All templates are centered at $(0,0)$.  In
every scene there are at most two squares and one triangle.  Each shape
is transformed with a random transformation to create a scene of 2D
points given by the object parts.  To match the evaluation of
\cite{kosiorek2019stacked}, all scenes are normalized so as the points
lie in $[-1,1]$ on both dimensions.  When creating the scene, we
select randomly (with probability 0.5) whether an object is going to
be present or not, but delete empty scenes.  A test set used for
evaluation is comprised of 450-460 non-empty scenes, based on 512
draws.

Additionally, we study how the methods compare when objects are
created from noisy templates.  We consider that the original templates
used for the creation of the images are corrupted with Gaussian noise
with standard deviation $\sigma$.  Once the templates are corrupted
with noise, a random transformation $\y_k$ is applied to obtain the
object $X_k$ of the scene.  As with the noise-free data, the objects
are normalized to lie in [-1,1] on both dimensions.

CCAE is trained by creating random batches of 128 scenes as described
above and optimizing the objective function in~\eqref{eq:CCAE_loss}.
The authors run CCAE for $300K$ epochs, and when the parameters of the
neural networks are trained, they use their model on the test dataset
to generate an estimation of which points belong to which capsule, and
where the estimated points are located in each scene.

The variational inference approach
allows us to model scenes where the points are corrupted with some
noise.  The annealing parameter $\beta$ controls the level of noise allowed in the model.  We use an annealing strategy to fit $\beta$, increasing it every time the ELBO has converged, up to a maximum value of $\beta_{max}=1$. We set
the hyperparameters of the model to $\bmu^g_0=\mathbf{0}$, $D^g_0 =
I_4$, $\lambda = 10^{4}$ and $a_{mnk} = \frac{1}{N}$.  We run
Algorithm~\ref{alg:variational_inference} with 5 different random
initializations of $R$ and select the solution with the best ELBO.
Similarly to~\citet{kosiorek2019stacked}, we incorporate a sparsity
constraint in our model, that forces every object to explain at least
two parts.  Once our algorithm has converged, for a given $k$ if any
$r_{mnk}>0.9$ and $\sum_m\sum_n r_{mnk} < 2$ it means that the model
has converged to a solution where object $k$ is assigned to less than
2 parts.  In these cases, we re-run
Algorithm~\ref{alg:variational_inference} with a new initialization of
$R$.  Notice that this is also related to the minimum basis size
necessary in the RANSAC approach for the types of transformations that
we are considering.

The implementation RANSAC 
(Algorithm~\ref{alg:geometric_hashing}) considers all matches between
the predicted and the scene points where the distance between them is
less than $0.1$. Among them, it selects the matching with minimum
distance between scene and predicted points.
For both the variational inference algorithm and the RANSAC algorithm,
a training dataset is not necessary.
These algorithms are applied directly to each test scene
and return a set of objects describing the points in the scene.
Unfortunately we do not have access to the code employed
by~\citet{hinton2018matrix}, so we have been unable to make
comparisons with it.

{\bf Evaluation:}
Three metrics are used to evaluate the performance of the different methods:
variation of information, adjusted Rand index and segmentation
accuracy.  They are based on partitions of the datapoints into those
associated with each object, and those that are missing.  Compared to
standard clustering evaluation metrics, some modifications are needed
to handle the missing objects. Details are provided in
supp.\ mat.\ \ref{sec:supp_eval}. We also use an average
scene accuracy metric, where a scene is correct if the method returns
the full original scene, and is incorrect otherwise.

\subsection{Parts-based face model} \label{sec:parts_face}

\begin{figure}
    \centering
    \begin{tabular}{cc}
        \includegraphics[scale=0.3]{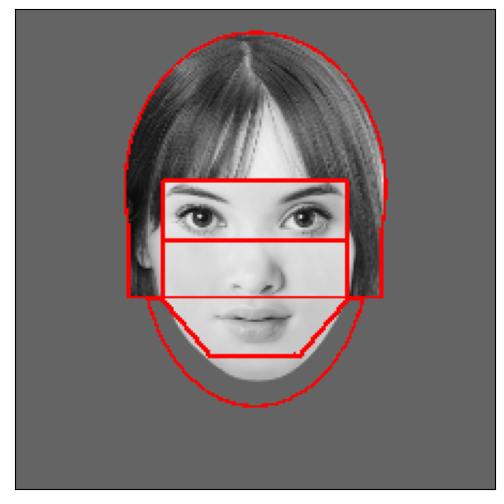} &
        \includegraphics[scale=0.3]{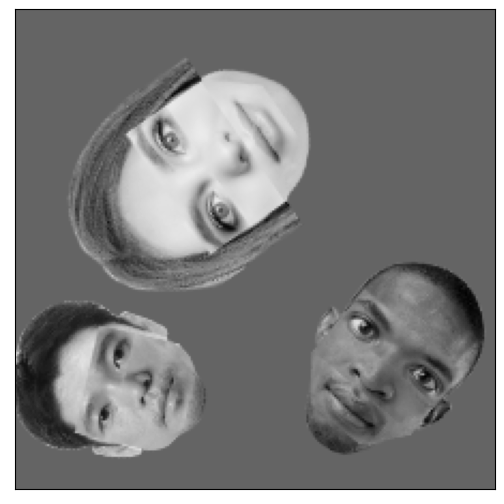}\\
        (a) & (b) \\
    \end{tabular}
\caption{(a) A synthetic face. The red lines indicate the
  areas of the 5 part types (i.e. hair, eyes, nose, mouth and
  jaw). (b) Example scene with 3 randomly transformed faces. \label{fig:face_example}}
\end{figure}

We have developed a novel hierarchical parts-based model for face
appearances. It is based on five parts, namely eyes, nose, mouth, hair
and forehead, and jaw (see Fig. \ref{fig:face_example}(a)). Each part
has a specified mask, and we have cropped the hair region to exclude
highly variable hairstyles. This decomposition is based on the
"PhotoFit Me" work and data of Prof.\ Graham Pike, see
\url{https://www.open.edu/openlearn/PhotoFitMe}.  For each part we
trained a probabilistic PCA (PPCA) model to reduce the dimensionality
of the raw pixels; the dimensionality is chosen so as to explain 95\%
of the variance. This resulted in dimensionalities of 24, 11, 12, 16
and 28 for the eyes, nose, mouth, jaw and hair parts respectively. We
then add a factor analysis (FA) model on top with latent variables
$\y^a_k$ to model the correlations of the PPCA coefficients across
parts. The dataset used (from PhotoFit Me) is balanced by gender
(female/male) and by race (Black/Asian/Caucasian), hence the
high-level factor analyser can model regularities across the parts,
e.g.\ wrt skin tone. $\x^a_{n}$ is predicted from $\y^a_k$ as 
$F^a_{kn} \y^a_{k} + \boldsymbol{m}^a_{kn}$ as in (\ref{eq:pxa_supp}).
  $\y^g_k$ would have an effect on the part appearance, e.g.\ by scaling
and rotation, but this can be removed by running the PPCA part
detectors on copies of the input image that have been rescaled and rotated.

The ``PhotoFit Me'' project utilizes 7 different part-images for each
gender/race group, for each of the five part types. As a result, we
generated $7^5$  synthetic faces for each group, by combining these
face-parts, which led to a total of $100,842$ faces. All faces were
centered on a $224\times224$ pixel canvas. For each synthetic face we
created an appearance vector $\x^a_n$, which consisted of the stacked
vectors from the 5 different principal component subspaces. Finally,
we created a balanced subset from the generated faces ($18,000$
images), which we used to train a FA model. We tuned the latent dimension of
this model by training it multiple times with a different number of
factors, and finally chose 12 factors, where a knee in the
reconstruction loss on the face data was observed on a validation set. 

Early work on PCA for faces--—"eigenfaces"
\citep{sirovich-kirby-87,turk-pentland-91}---used a global basis as
opposed to a parts-based representation. \citet{rao-ballard-99} made
use of a hierarchical factor analysis model, but used it to model
extended edges in natural image patches rather than facial
parts. \citet{ross-zemel-06} developed a ``multiple cause factor
analysis'' (MCFA) and applied it to faces; in contrast to our work
this did not have a higher-level factor analyser to model correlations
between parts, but it did allow variability in the masks of the parts.

\citet{kosiorek2019stacked} developed a Part Capsule Autoencoder
(PCAE) to learn parts from images, and applied it to MNIST images.
Each PCAE part is a template which can undergo an affine
transformation, and it has "special features" that were used to encode
the colour of the part. Thus the PCAE parts are less flexible than our
facial parts---they only model global colour changes of the part, and
not the rich variability that can be encoded with the PCA model. Also,
we have found that the parts detected by PCAE are not equivariant to
rotation. Figure \ref{fig:supp_PCAE_equiv} shows the PCAE part
decompositions inferred for different angles of rotation of a 4
digit---notice e.g.\ in panel (a) how the part coloured white maps to different
parts of the 4 for $45^{\circ}$-$180^{\circ}$ and $225^{\circ}$-$0^{\circ}$.

To evaluate our inference algorithm we generated $224\times224$ pixel
scenes of faces. These consisted of 2, 3, 4 or 5
randomly selected faces from a balanced test-set of $7,614$ synthetic
faces, which were transformed with random similarity
transformations. The face-objects were randomly scaled down by a
minimum of 50\% and were also randomly translated and rotated, with
the constraint that all the faces fit the scene and did not overlap
each other. An example of such a scene is shown in Fig.\
\ref{fig:face_example}(b), and further examples are shown in the supp.\ mat.\
Figure \ref{fig:face_inference_examples}.
Afterwards, these two constraints were dropped to test the
ability of our model to perform inference with occluded 
parts, see Figure \ref{fig:face_inference_examples}(e) for
an example. In our experiments we assume that the face parts
are detected accurately, i.e.\ as generated.

In the case of facial parts---and parts with both geometric and
appearance features in general---it only makes sense to assign the
observed parts $\x_m$ to template parts $\x_{kn}$ of the same type (e.g. an
observed ``nose'' part should be assigned only to a template ``nose''
part). We assume that this information is known, since the size of the
appearance vector of each part-type is unique. Thus it no
longer makes sense to initialize the assignment matrix uniformly for
all entries, but rather only for the entries that correspond to
templates of the observed part's type. Consequently,
\eqref{eq:log_rho} is only utilized for $m, n$ pairs of the same
type. Similarly to the constellation experiments, we initialized the
assignment matrix 5 times and selected the solution with the largest
ELBO.
  
In the experiments we evaluated the part assignment accuracy of the
algorithms. In a given scene, the assignment is considered correct if
all the observed parts have been correctly assigned to their
corresponding template parts with the highest probability. In
order to evaluate the prediction of the appearance features, we
measured the root mean square error (RMSE) between the input and
generated scenes. 


\section{Results}
We present results for the constellations and faces experiments in
sections  \ref{sec:const_res} and  \ref{sec:face_res}
respectively.

\subsection{Constellation Experiments} \label{sec:const_res}
In Table~\ref{t:main_results} we show a comparison between CCAE, the
variational inference method with a Gaussian mixture
prior~\eqref{eq:p_Z} (GCM-GMM), with a prior over permutation
matrices~\eqref{eq:p_Z_true} (GCM-DS), and the RANSAC
approach~\footnote{Code at: https://github.com/anazabal/GenerativeCapsules}.
For GCM-GMM and GCM-DS we show the results where the
initial  $\beta = 0.05$,  
relegating a comparison across different initializations of $\beta$ to the
supp.\ mat.\ \ref{sec:lambda_study} and \ref{sec:GTmissing}.
Scenes without noise and with noise levels of $\sigma = 0.1, \; 0.25$
are considered.

\begin{table}[t]
\centering
\caption{Comparison between the different methods. For SA, ARI and Scene Accuracy the higher the better. For VI the lower the better. Different levels of Gaussian noise with standard deviation $\sigma$ are considered.}
\begin{tabular}{|c|c|c|c|c|}
\hline
Metric & Model & $\sigma$=0 & $\sigma$=0.1 & $\sigma$=0.25 \\
\hline
\multirow{4}{*}{SA $\uparrow$} & CCAE & 0.828 & 0.754 & 0.623\\
& GCM-GMM & 0.753 &     0.757 & 0.744\\
& GCM-DS & 0.899 &      0.882 & 0.785\\
& RANSAC & 1 & 0.992 & 0.965\\\hline
\multirow{4}{*}{ARI $\uparrow$} & CCAE & 0.599 & 0.484 & 0.248\\
& GCM-GMM & 0.586 &     0.572 & 0.447\\
& GCM-DS & 0.740 &      0.699 & 0.498\\
& RANSAC & 1 & 0.979 & 0.914\\\hline
\multirow{4}{*}{VI $\downarrow$} & CCAE & 0.481 & 0.689 & 0.988\\
& GCM-GMM & 0.478 &     0.502  & 0.677\\
& GCM-DS &      0.299 & 0.359 & 0.659\\
& RANSAC & 0 & 0.034  & 0.135\\\hline
& CCAE & 0.365 & 0.138 & 0.033\\
Scene & GCM-GMM & 0.179 &       0.173  & 0.132\\
{Acc $\uparrow$} & GCM-DS &     0.664 & 0.603  & 0.377\\
& RANSAC & 1 & 0.961 & 0.843\\\hline
\end{tabular}
\label{t:main_results}
\end{table}
  
We see that GCM-DS improves over CCAE and GCM-GMM in all of the
metrics, with GCM-GMM being comparable to CCAE.  Interestingly, for
the noise-free scenarios, the RANSAC method achieves a perfect score
for all of the metrics.  Since there is no noise on the observations
and the method searches over all possible solutions of $\y_k$, it can
find the correct solution for any configuration of geometric shapes in
a scene.  For the noisy scenarios, all the methods degrade as $\sigma$
increases.  However, the relative performance between them remains the
same, with RANSAC performing the best, followed by GCM-DS and then
GCM-GMM. Example inferences are shown and discussed in
supp.\ mat. \ref{sec:ccae_gcmds_inferences} and
\ref{sec:noisy_examples}.

\subsection{Face Experiments} \label{sec:face_res}
Firstly, the VI algorithm was evaluated on scenes of multiple,
randomly selected and transformed faces~\footnote{Code at: https://github.com/tsagkas/capsules}.
For scenes with 2, 3, 4 and 5
faces, the assignment accuracy was 100\%, 100\%, 99.2\% and 93.7\%
respectively (based on 250 scenes per experiment). 
RANSAC gave 100\% accurate assignments in all 4
cases. This is to be expected, since from each part the pose of the
whole can be predicted accurately. However, RANSAC's ability to infer the
appearance of the faces proved to be limited. More specifically, in
  250 instances uniformly distributed across scenes of 2, 3, 4 and 5
  faces, the VI algorithm had RMSE of $0.036\pm0.004$, while
  RANSAC scored $0.052\pm0.006$, with consistently higher error on \emph{all}
  scenes. This is illustrated in the examples of Figure
  \ref{fig:face_inference_examples}, where it is clear that RANSAC is
  less accurate in capturing key facial characteristics.
If inference for $\y^a_k$ is run as a post-processing step for
  RANSAC using all detected parts in an object, this difference disappears.
 
The supp.\ mat. contains a movie showing the fitting of the
models to the data. It is not possible for us to compare results with
the SCAE algorithm on the faces data, as the PCAE model used is
not rich enough to model PCA subspaces.

Secondly, we evaluated the ability of our algorithm to perform
inference in scenes where some parts have been occluded, either by
overlapping with other faces or by extending out of the scene. In 250
scenes with 3 partially occluded faces, both the VI and RANSAC
algorithms were fully successful in assigning the observed parts to
the corresponding template accurately. See Figure
\ref{fig:face_inference_examples}(e) for an example.

\section{Discussion}
Above we have described a principled generative capsules model. It
leads directly to a variational inference algorithm which can handle
either a mixture formulation or incorporation of the DS matrix
constraint.  We have also shown that this formulation outperforms the
CCAE of \citet{kosiorek2019stacked} on the constellation data
generator that they used.

In our experiments RANSAC was shown to often be an effective
  alternative to variational inference (VI). This is particularly the
  case when the basis in RANSAC is highly informative about the
  object.  RANSAC provides an interesting alternative to routing-by-agreement,
  although its sampling-based inference is less amenable to
  neural-network style parallel implementations than VI.

In this paper we have not discussed the learning of the geometric model
(i.e.\ the $F^g_{kn}$ matrices), but in all such work, inference for
$q(Z,Y)$ forms an inner loop, with model learning as an outer loop.
Also, the interpretable nature of the model means that we can exploit
knowledge where it is available; in contrast the SCAE's
opaque inference networks do not allow this.

\begin{acknowledgements} 
%
This work was supported in part by The Alan Turing Institute under
EPSRC grant EP/N510129/1.
\end{acknowledgements}


\bibliography{references}

\begin{thebibliography}{27}
\providecommand{\natexlab}[1]{#1}
\providecommand{\url}[1]{\texttt{#1}}
\expandafter\ifx\csname urlstyle\endcsname\relax
  \providecommand{\doi}[1]{doi: #1}\else
  \providecommand{\doi}{doi: \begingroup \urlstyle{rm}\Url}\fi

\bibitem[Biederman(1987)]{biederman-87}
I.~Biederman.
\newblock {Recognition-by-components: A theory of human image understanding}.
\newblock \emph{{Psychological Review}}, 94:\penalty0 115--147, 1987.

\bibitem[Bishop(2006)]{bishop2006pattern}
C.~M. Bishop.
\newblock \emph{{Pattern Recognition and Machine Learning}}.
\newblock Springer, 2006.

\bibitem[Fischler and Bolles(1981)]{fischler1981random}
M.~A. Fischler and R.~C. Bolles.
\newblock Random sample consensus: a paradigm for model fitting with
  applications to image analysis and automated cartography.
\newblock \emph{Communications of the ACM}, 24\penalty0 (6):\penalty0 381--395,
  1981.

\bibitem[Hinton et~al.(2000)Hinton, Ghahramani, and
  Teh]{hinton-ghahramani-teh-00}
G.~E. Hinton, Z.~Ghahramani, and Y.~W. Teh.
\newblock {Learning to Parse Images}.
\newblock In S.~A. Solla, T.~K. Leen, and K.-R. M\"{u}ller, editors,
  \emph{{Advances in Neural Information Processing Systems 12}}, pages
  463--469. MIT Press, Cambridge, MA, 2000.

\bibitem[Hinton et~al.(2018)Hinton, Sabour, and Frosst]{hinton2018matrix}
G.~E. Hinton, S.~Sabour, and N.~Frosst.
\newblock {Matrix capsules with EM routing}.
\newblock In \emph{{International Conference on Learning Representations}},
  2018.

\bibitem[Hubert and Arabie(1985)]{hubert1985comparing}
L.~Hubert and P.~Arabie.
\newblock Comparing partitions.
\newblock \emph{Journal of Classification}, 2\penalty0 (1):\penalty0 193--218,
  1985.

\bibitem[Kosiorek et~al.(2019)Kosiorek, Sabour, Teh, and
  Hinton]{kosiorek2019stacked}
A.~Kosiorek, S.~Sabour, Y.~W. Teh, and G.~E. Hinton.
\newblock Stacked capsule autoencoders.
\newblock In \emph{{Advances in Neural Information Processing Systems}}, pages
  15512--15522, 2019.

\bibitem[Lee et~al.(2019)Lee, Lee, Kim, Kosiorek, Choi, and
  Teh]{lee-lee-kosiorek-choi-teh-19}
J.~Lee, Y.~Lee, J.~Kim, A.~R. Kosiorek, S.~Choi, and Y.-W. Teh.
\newblock {Set Transformer: A Framework for Attention-based
  Permutation-invariant Neural Networks}.
\newblock In \emph{{Proc. of the 36th International Conference on Machine
  Learning}}, pages 3744--3753, 2019.

\bibitem[Li and Zhu(2019)]{li-zhu-19}
Y.~Li and X.~Zhu.
\newblock {Capsule Generative Models}.
\newblock In I.~V. Tetko et~al., editors, \emph{{ICANN}}, pages 281--295. LNCS
  11727, 2019.

\bibitem[Li et~al.(2020)Li, Zhu, Naud, and Xi]{li-zhu-naud-xi-20}
Y.~Li, X.~Zhu, R.~Naud, and P.~Xi.
\newblock {Capsule Deep Generative Model That Forms Parse Trees}.
\newblock In \emph{2020 International Joint Conference on Neural Networks
  (IJCNN)}. IEEE, 2020.

\bibitem[Meil{\u{a}}(2003)]{meilua2003comparing}
M.~Meil{\u{a}}.
\newblock Comparing clusterings by the variation of information.
\newblock In \emph{Learning Theory and Kernel Machines}, pages 173--187.
  Springer, 2003.

\bibitem[Mena et~al.(2020)Mena, Varol, Nejatbakhsh, Yemini, and
  Paninski]{mena2020sinkhorn}
Gonzalo Mena, Erdem Varol, Amin Nejatbakhsh, Eviatar Yemini, and Liam Paninski.
\newblock Sinkhorn permutation variational marginal inference.
\newblock In \emph{Symposium on Advances in Approximate Bayesian Inference},
  pages 1--9. PMLR, 2020.

\bibitem[Powell and Smith(2019)]{powell2019computing}
B.~Powell and P.~A. Smith.
\newblock Computing expectations and marginal likelihoods for permutations.
\newblock \emph{Computational Statistics}, pages 1--21, 2019.

\bibitem[Rand(1971)]{rand1971objective}
W.~M. Rand.
\newblock Objective criteria for the evaluation of clustering methods.
\newblock \emph{Journal of the American Statistical Association}, 66\penalty0
  (336):\penalty0 846--850, 1971.

\bibitem[Rao and Ballard(1999)]{rao-ballard-99}
R.~P.~N. Rao and D.~H. Ballard.
\newblock {Predictive coding in the visual cortex: a functional interpretation
  of some extra-classical receptive-field effects}.
\newblock \emph{Nature Neurosci.}, 2(1):\penalty0 79--87, 1999.

\bibitem[Revow et~al.(1996)Revow, Williams, and
  Hinton]{revow-williams-hinton-96}
M.~Revow, C.~K.~I. Williams, and G.~E. Hinton.
\newblock {Using Generative Models for Handwritten Digit Recognition}.
\newblock \emph{IEEE Trans. on Pattern Analysis and Machine Intelligence},
  18(6):\penalty0 592--606, 1996.

\bibitem[Ross and Zemel(2006)]{ross-zemel-06}
D.~Ross and R.~Zemel.
\newblock {Learning parts-based representations of data}.
\newblock \emph{Journal of Machine Learning Research}, 7:\penalty0 2369--2397,
  2006.

\bibitem[Sabour et~al.(2017)Sabour, Frosst, and Hinton]{sabour2017dynamic}
S.~Sabour, N.~Frosst, and G.E. Hinton.
\newblock Dynamic routing between capsules.
\newblock In \emph{{Advances in Neural Information Processing Systems}}, pages
  3856--3866, 2017.

\bibitem[Sinkhorn and Knopp(1967)]{sinkhorn1967concerning}
R.~Sinkhorn and P.~Knopp.
\newblock Concerning nonnegative matrices and doubly stochastic matrices.
\newblock \emph{Pacific Journal of Mathematics}, 21\penalty0 (2):\penalty0
  343--348, 1967.

\bibitem[Sirovich and Kirby(1987)]{sirovich-kirby-87}
L.~Sirovich and M.~Kirby.
\newblock {Low-dimensional Procedure for the Characterization of Human Faces}.
\newblock \emph{Journal of the Optical Society of America A}, 4(3):\penalty0
  519--524, 1987.

\bibitem[Smith et~al.(2021)Smith, Schut, Gal, and van~der
  Wilk]{smith-schut-gal-vanderwilk-21}
L.~Smith, L.~Schut, Y.~Gal, and M.~van~der Wilk.
\newblock {Capsule Networks---A Generative Probabilistic Perspective}, 2021.
\newblock https://arxiv.org/pdf/2004.03553.pdf.

\bibitem[Storkey and Williams(2003)]{storkey-williams-03}
A.~J. Storkey and C.~K.~I. Williams.
\newblock {Image Modelling with Position-Encoding Dynamic Trees}.
\newblock \emph{{IEEE Trans Pattern Analysis and Machine Intelligence}},
  25(7):\penalty0 859--871, 2003.

\bibitem[Torr(1998)]{torr-98}
P.~H.~S. Torr.
\newblock {Geometric Motion Segmentation and Model Selection}.
\newblock \emph{Philosophical Trans. of the Royal Society A}, 356:\penalty0
  1321--1340, 1998.

\bibitem[Turk and Pentland(1991)]{turk-pentland-91}
M.~Turk and A.~Pentland.
\newblock {Eigenfaces for Recognition}.
\newblock \emph{Journal of Cognitive Neuroscience}, 3(1):\penalty0 71--86,
  1991.

\bibitem[Vincent and Lagani\`{e}re(2001)]{vincent-laganiere-01}
E.~Vincent and R.~Lagani\`{e}re.
\newblock Detecting planar homographies in an image pair.
\newblock In \emph{Proceedings of the 2nd International Symposium on Image and
  Signal Processing and Analysis (ISPA)}, 2001.

\bibitem[Williams et~al.(2019)Williams, Nash, and
  Nazabal]{williams-nash-nazabal-19}
C.~K.~I. Williams, C.~Nash, and A.~Nazabal.
\newblock {Autoencoders and Probabilistic Inference with Missing Data: An Exact
  Solution for The Factor Analysis Case}, 2019.
\newblock arXiv:1801.03851.

\bibitem[Wolfson and Rigoutsos(1997)]{wolfson1997geometric}
H.~J. Wolfson and I.~Rigoutsos.
\newblock Geometric hashing: An overview.
\newblock \emph{IEEE Computational Science and Engineering}, 4\penalty0
  (4):\penalty0 10--21, 1997.

\end{thebibliography}

\clearpage
\newpage

\appendix
\begin{center}
{\LARGE Supplementary Material for \emph{Inference for
  Generative Capsule Models}}
\end{center}

\section{Details for Variational Inference}\label{sec:appVI}

The Sinkhorn-Knopp algorithm is given in Algorithm \ref{alg:sinkhorn_knopp}.

\begin{algorithm}[h]
\caption{Sinkhorn-Knopp algorithm}
\label{alg:sinkhorn_knopp}
\begin{algorithmic}[1]
\Procedure{SinkhornKnopp}{$X$}
	\While{$X$ not doubly stochastic}
		\State Normalize rows of $X$: $x_{ij} = \frac{x_{ij}}{\sum_{j} x_{ij}}$, $\forall{i}$
		\State Normalize columns of $X$: $x_{ij} = \frac{x_{ij}}{\sum_{i} x_{ij}}$, $\forall{j}$
	\EndWhile
	\Return $X$
\EndProcedure
\end{algorithmic}
\end{algorithm}

The evidence lower bound (ELBO) 
$L(q)$ for this model is decomposed in three terms:
\begin{align}\label{eq:ELBO}
L(q) &=\bE_q[\log{p(X|Y,Z)}] \\\nonumber
&- KL(q(Y)||p(Y)) - KL(q(Z)||p(Z)),
\end{align}
with $KL(q||p)$ being the Kullback-Leibler divergence between
distributions $q$ and $p$. 
The first term indicates how well the generative model $p(X|Y,Z)$ fits the observations under our variational model $q(Y,Z)$:
\begin{align}\label{eq:E_q_X}
&\bE_q[\log{p(X|Y,Z)}] = \nonumber \\
&-\sum_{m=1}^M\sum_{k=1}^K\sum_{n=1}^{N_k} r_{mnk} \Big[
    \frac{d_{kn}}{2} \log{2\pi} +\frac{1}{2}\log{|\beta^{-1} D_{kn}|} \Big.  \nonumber\\
&+ (\x_{m} - F_{kn}\bmu_{k}- \boldsymbol{m}_{kn})^T D_{kn}^{-1}(\x_{m} - F_{kn}\bmu_{k} \boldsymbol{m}_{kn}) \nonumber\\[7pt]
& \hspace{3cm} +
      \Big. \text{trace}(F_{kn}^TD_{kn}^{-1}F_{kn}\Lambda_{k}^{-1}) \Big] .
\end{align}

The Kullback-Leibler divergence between the two Gaussian distributions $q(Y)$ and $p(Y)$ in our model has the following expression:
\begin{gather}\nonumber  
KL(q(Y)||p(Y)) = \frac{1}{2} \sum_{k=1}^K \left( 
\text{trace}(D^{-1}_0 \Lambda^{-1}_k ) - d_k + \right.\\
 \left. (\bmu_k -\bmu_0)^T D_0^{-1}(\bmu_k - \bmu_0) +
\log{|D_{0}|} + \log{|\Lambda_k|} \right),
\end{gather}
where $d_k$ is the dimensionality of $\y_k$.

The expression for $KL(q(Z)||p(Z))$ is given by
\begin{align}\label{eq:KL_Z}
KL(q(Z)||p(Z)) &= \sum_{m=1}^{N}\sum_{k=1}^K\sum_{n=1}^{N_k}
r_{mnk}\log{\frac{r_{mnk}}{a_{mnk}}} .
\end{align}

The expectation term in~\eqref{eq:log_rho} is:
\begin{align}
& \bE_{\y_{k}}[(\x_{m}-F_{kn}\y_{k}-\boldsymbol{m}_{kn})^T D_{kn}^{-1} (\x_{m}-F_{kn}\y_{k}-\boldsymbol{m}_{kn})] \nonumber\\
& =(\x_{m} - F_{kn}\bmu_{k}- \boldsymbol{m}_{kn})^T D_{kn}^{-1}(\x_{m} -
F_{kn}\bmu_{k} - \boldsymbol{m}_{kn})  \nonumber\\
& \hspace{3cm}   + \text{trace}(F_{kn}^TD_{kn}^{-1}F_{kn}\Lambda_{k}^{-1}).
\end{align}

\begin{algorithm}[t]
\caption{RANSAC approach \label{alg:geometric_hashing}}
\begin{algorithmic}[1]
\State $T$: $K$ templates of the scene
\State $B_k$: base matrix for template $T_k$
\State $X$: $M$ points of the scene
\State $out = []$
        \For{$\x_i \in X$}
                \For{$\x_j \in X \setminus x_i$}
                        \State $\x_{ij} = \text{Vectorize}(\x_i,\x_j)$
                        \For{$k = 1:K$}
                                \State $\hat{\y}_k = B_k^{-1}\x_{ij}$
                                \State $T_k \xrightarrow{\hat{\y}_k} \hat{X}_k$
                                \If{ \text{SubsetMatch}($\hat{X}_k, X)$}
                                        \State Add $(T_k,\hat{\y}_k,\hat{X}_k)$ to $out$
                                \EndIf
                        \EndFor
                \EndFor
        \EndFor
\Return $out$
\end{algorithmic}
\end{algorithm}

\section{Details of RANSAC for the constellations data}
\label{sec:supp_RANSAC}
A summary of the algorithm is given in Algorithm~\ref{alg:geometric_hashing}.
  
Assume we have chosen parts $n_1$ and $n_2$ as the basis for object
$k$, and that we have selected datapoints $\x_i$ and $\x_j$ as their
hypothesized counterparts in the image.
Let $\x_{ij}$ be the vector obtained by stacking $\x_i$ and $\x_j$, and $B_k$ be the $4\times 4$ square matrix obtained by stacking $F_{k n_1}$ and $F_{k n_2}$.
Then $\hat{\y}_k = B_{k}^{-1}\x_{ij}$.
Finally, $\text{SubsetMatch}(\hat{X}_k, X)$ selects those points in $X_k$ that are close to $X$ with a given tolerance and add them to the output. Among them, the solution is given by the one that minimizes $\sum_{n=1}^{N_k}(\hat{x}_{nk} - x_{nk})^2$.

The above algorithm chooses a specific basis for each object, but one
can consider all possible bases for each object.  It is then efficient
to use a hash table to store the predictions for each part, as used
in Geometric Hashing~\citep{wolfson1997geometric}.  Geometric Hashing
is traditionally employed in computer vision to match geometric
features against previously defined models of such features.  This
technique works well with partially occluded objects, and is
computationally efficient if the basis dimension is low.

\section{Evaluation metrics for the constellations data} \label{sec:supp_eval}
In a given scene $X$ there are $M$ points, but we know that there are
$N \ge M$ possible points that can be produced from all of the
templates.  Assume that $K' \le K$ templates are active in this scene.
Then the points in the scene are labelled with indices $1, \ldots,
K'$, and we assign the missing points index $0$.  Denote the ground
truth partition as $V = \{ V_0, V_1, \ldots V_{K'} \}$.  An
alternative partition output by one of the algorithms is denoted by
$\hat{V} = \{ \hat{V}_0, \hat{V}_1, \ldots \hat{V}_{\hat{K}'} \}$.
The predicted partition $\hat{V}$ may instantiate objects or points
that were in fact missing, thus it is important to handle the missing
data properly.

In Information Theory, the \textbf{variation of information}
(VI)~\citep{meilua2003comparing} is a measure of the distance between
two partitions of elements (or clusterings).  For a given set of
elements, the variation of information between two partitions $V$ and
$\hat{V}$, where $N = \sum_i |V_i| = \sum_j |\hat{V}_j|$ is defined
as:
\begin{equation}
VI(V,\hat{V}) = -\sum_{i,j} r_{ij}\left[\log{\frac{r_{ij}}{p_i}} + \log{\frac{r_{ij}}{q_j}} \right]
\end{equation}
where $r_{ij} = \frac{|V_i \cap \hat{V}_j|}{N}$, $p_i =
\frac{|V_i|}{N}$ and $q_j = \frac{|\hat{V}_j|}{N}$.  In our
experiments we report the average variation of information of the
scenes in the dataset.

The \textbf{Rand index}~\citep{rand1971objective} is another measure
of similarity between two data clusterings.  This metric takes pairs
of elements and evaluates whether they do or do not belong to the same
subsets in the partitions $V$ and $\hat{V}$
\begin{equation}
RI = \frac{TP + TN}{TP +TN +FP +FN},
\end{equation}
where TP are the true positives, TN the true negatives, FP the false
positives and FN the false negatives.  The Rand index takes on values
between 0 and 1.  We use instead the \textbf{adjusted Rand index}
(ARI) \citep{hubert1985comparing}, the corrected-for-chance version of
the Rand index.  It uses the expected similarity of all pair-wise
comparisons between clusterings specified by a random model as a
baseline to correct for assignments produced by chance.  Unlike the
Rand index, the adjusted Rand index can return negative values if the
index is less than the expected value.  In our experiments, we compute
the average adjusted Rand index of the scenes in our dataset.

The \textbf{segmentation accuracy} (SA) is based on obtaining the
maximum bipartite matching between $V$ and $\hat{V}$, and was used by
\citet{kosiorek2019stacked} to evaluate the performance of CCAE.  For
each set $V_i$ in $V$ and set $\hat{V}_j$ in $\hat{V}$, there is an
edge $w_{ij}$ with the weight being the number of common elements in
both sets.  Let $W(V,\hat{V})$ be the overall weight of the maximum
matching between $V$ and $\hat{V}$.  Then we define the average
segmentation accuracy as:
\begin{equation}
SA = \sum_{i=1}^I  \frac{W(V_i,\hat{V}_i)}{W(V_i,V_i)} = \frac{1}{N}\sum_{i=1}^I W(V_i,\hat{V}_i),
\end{equation}
where $I$ is the number of scenes.
Notice that $W(V_i,V_i)$ represents a perfect assignment of the ground
truth, both the observed and missing subsets, and thus $W(V_i,V_i) =
N$.

There are some differences on how we compute the SA metric compared
to~\cite{kosiorek2019stacked}.  First, they do not consider the
missing points as part of their ground truth, but as we argued above
this is necessary.  They evaluate the segmentation accuracy in terms
of the observed points in the ground truth, disregarding possible
points that were missing in the ground truth but predicted as observed
in $\hat{V}$.  Second, they average the segmentation accuracy across
scenes as
\begin{equation}
  SA =   \frac{\sum_{i=1}^I W(V_i,\hat{V}_i)}{\sum_{i=1}^I W(V_i,V_i)} .
\label{eq:SAkos}  
\end{equation}
For them, $W(V_i,V_i) = M_i$, where $M_i$ is the number of points
present in a scene.  In our case, both averaging formulae are
equivalent since our $W(V_i,V_i)$ is the same across scenes.

\section{Failures of rotation equivariance for SCAE} \label{sec:supp_PCAE}
We trained the SCAE model with digit ``4'' images from the training
set of the MNIST dataset\footnote{http://yann.lecun.com/exdb/mnist .},
after they had been uniformly
rotated by up to $360^\circ$ and uniformly translated by up to 6
pixels on the x and y axes. Since we used a single class in the
dataset we altered SCAE's architecture to use only a single object
capsule.

We repeated the training of SCAE multiple times for 8K epochs, and
collected distinct sets of learned $11\times11$ parts that the digit
``4'' can be composed into (see
Fig. \ref{fig:supp_templates}). Afterwards, we evaluated PCAE's
ability to detect these parts in MNIST digit ``4'' images that had
been rotated by multiples of $45^\circ$. Then, we measured the average
inferred pose of each part at each of the different angles of
rotations. Our results indicate that the PCAE model is not equivariant
to rotations. This is apparent from Fig. \ref{fig:supp_PCAE_equiv},
where the learned parts from part-sets 2 and 3 of
Fig. \ref{fig:supp_templates} are inconsistently assigned to the
regions of the digit-object, depending on the angle of rotation. We
hypothesize that this phenomenon stems from the fact that PCAE seems
to generate either parts that are characterized by an intrinsic
symmetry --and thus their pose is ambiguous-- (e.g. template 3 of
column (a) in Fig. \ref{fig:supp_templates}) or pairs of parts that
are transformed versions of themselves, and thus can be used
interchangeably (e.g. templates 1 and 3 of part-set 3 in
Fig. \ref{fig:supp_templates}). This leads to identifiability issues,
where the object can be decomposed into its parts in numerous ways.

\begin{figure}[t]
    \centering
    \caption{Each column corresponds to a part-set of learned templates parts of size 3, 4, 4 and 6 respectively.}
    \begingroup
    \setlength{\tabcolsep}{1pt} 
    \renewcommand{\arraystretch}{1} 
    \begin{tabular}{ccccc}
        \rotatebox[origin=c]{90}{template 1} & 
        \includegraphics[align=c,scale=0.18]{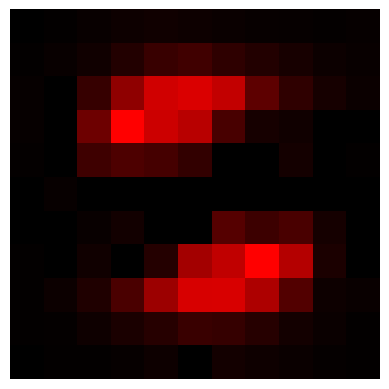} & 
        \includegraphics[align=c,scale=0.18]{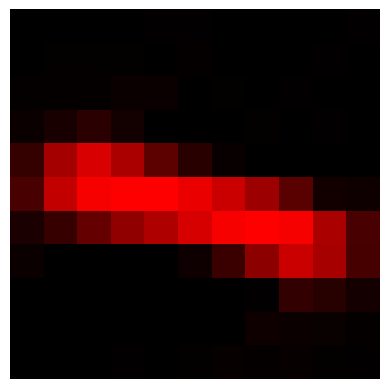} & 
        \includegraphics[align=c,scale=0.18]{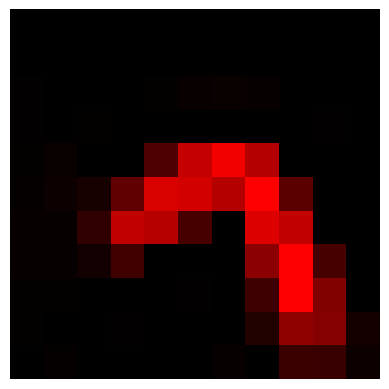} & 
        \includegraphics[align=c,scale=0.18]{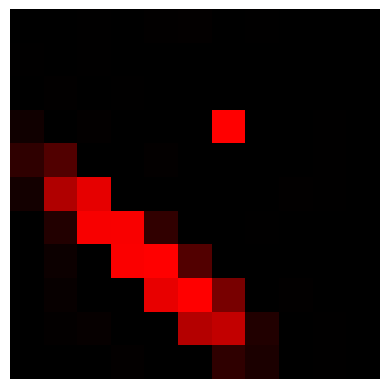} \\ 
        
        \rotatebox[origin=c]{90}{template 2} & 
        \includegraphics[align=c,scale=0.18]{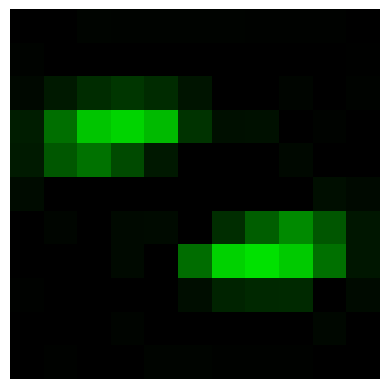} & 
        \includegraphics[align=c,scale=0.18]{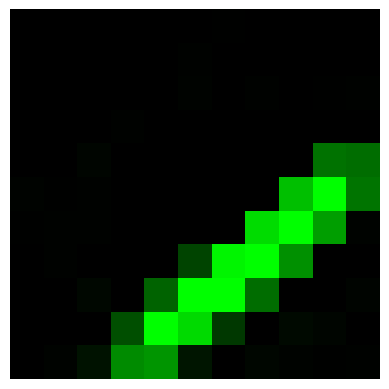} & 
        \includegraphics[align=c,scale=0.18]{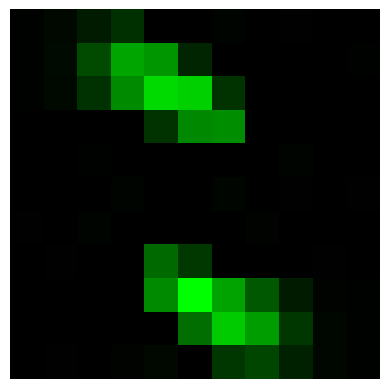} & 
        \includegraphics[align=c,scale=0.18]{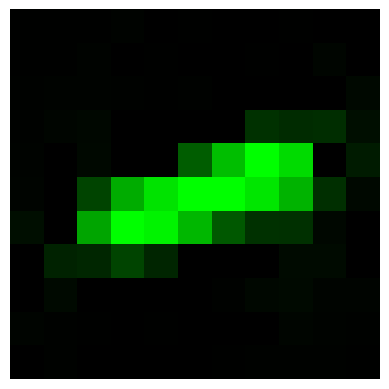} \\ 
        
        \rotatebox[origin=c]{90}{template 3} & 
        \includegraphics[align=c,scale=0.18]{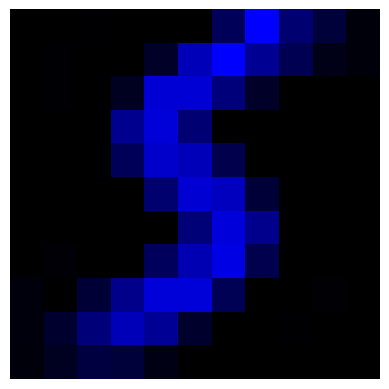} & 
        \includegraphics[align=c,scale=0.18]{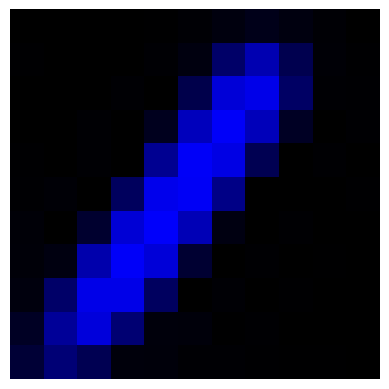} & 
        \includegraphics[align=c,scale=0.18]{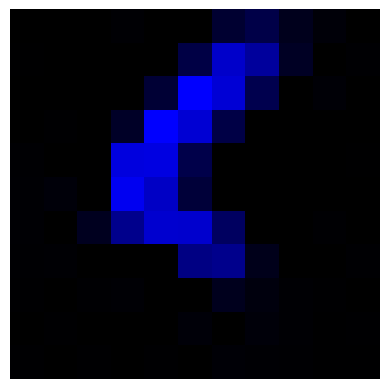} & 
        \includegraphics[align=c,scale=0.18]{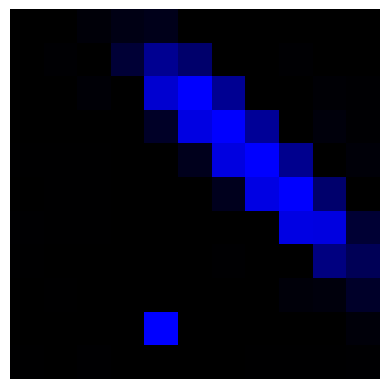} \\ 
        
         \rotatebox[origin=c]{90}{template 4} & 
        & 
        \includegraphics[align=c,scale=0.18]{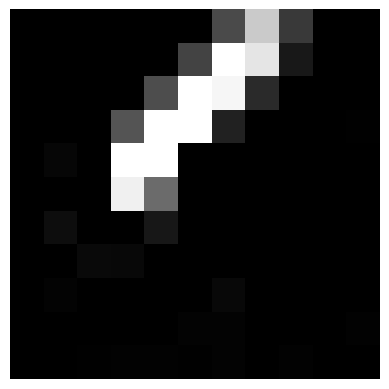} & 
        \includegraphics[align=c,scale=0.18]{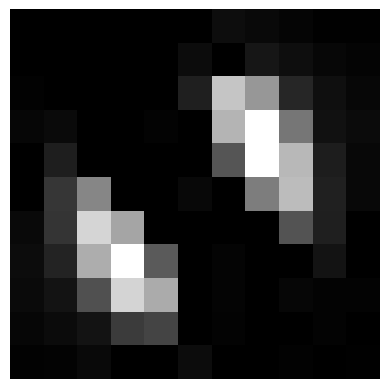} & 
        \includegraphics[align=c,scale=0.18]{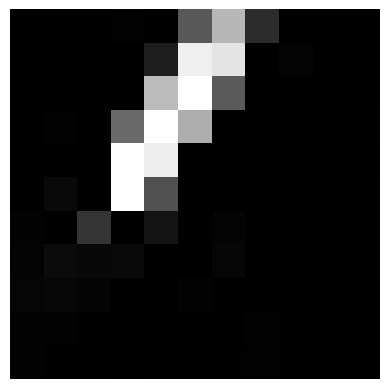} \\  
        
         \rotatebox[origin=c]{90}{template 5} & 
        & & & 
        \includegraphics[align=c,scale=0.18]{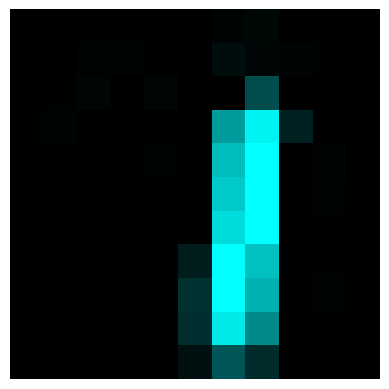} \\       
        
        \rotatebox[origin=c]{90}{template 6} & 
        & & & 
        \includegraphics[align=c,scale=0.18]{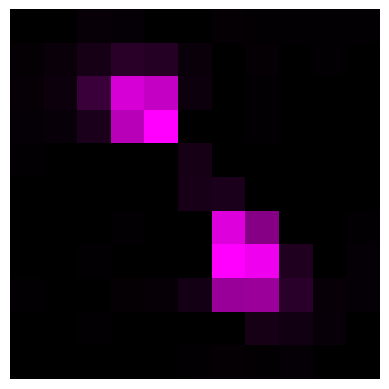} \\       
        
        &part-set 1&part-set 2&part-set 3&part-set 4\\
    \end{tabular}
    \endgroup

    \label{fig:supp_templates}
\end{figure}

\section{Different initializations of $\beta$}\label{sec:lambda_study}

In this section, we show the effect of varying the initial value of
$\beta$ for the variational inference method with a Gaussian mixture
prior ~\eqref{eq:p_Z} (GCM-GMM) and with a prior over permutation
matrices ~\eqref{eq:p_Z_true} (GCM-DS). We have tested 6 different
values of $\beta: 0.005, \; 0.01, \; 0.05, \; 0.1, \; 0.2, \; 0.5$.
In Tables~\ref{t:supp_results_sigma0}, \ref{t:supp_results_sigma01} and
\ref{t:supp_results_sigma025} we show the results for all metrics,
methods and initial values of $\beta$ for $\sigma=0$, $\sigma=0.1$,
  and $\sigma=0.25$ resp.

We can observe that no matter the initialization of $\beta$, GCM-DS is always better than CCAE and GCM-GMM. As we increase the initial $\beta$, GCM-DS performs better across all the metrics. We have found that the performance of GCM-DS and GCM-GMM degrades with $\beta > 0.1$.

We have conducted paired t-tests between CCAE and GCM-GMM, GCM-DS and
RANSAC on the three clustering metrics for $\sigma = 0$ and initial
$\beta = 0.05$. The differences between CCAE and GCM-DS
are statistically significant with p-values less than $10^{-7}$, and
between CCAE and RANSAC with p-values less than $10^{-28}$.
For CCAE and GCM-GMM the differences are not statistically
significant.

\begin{table}[h]
\centering
\caption{Different initializations of $\beta$. For SA, ARI and Scene
  Accuracy the higher the better. For VI the lower the better. Noise-free case.}
{\small
\begin{tabular}{|l|c|c|c|c|}
\hline
Model & SA $\uparrow$ & ARI $\uparrow$ & VI $\downarrow$ & Sc. Acc. $\uparrow$ \\
\hline
CCAE                   &          0.828 &     0.599 &    0.481 &           0.365 \\\hline
GCM-DS $\beta=0.005$    &          0.869 &     0.667 &    0.350 &           0.581 \\
GCM-DS $\beta=0.01$   &          0.876 &     0.678 &    0.328 &           0.624 \\
GCM-DS $\beta=0.05$   &          0.899 &     0.740 &    \bf 0.299 &          \bf 0.664 \\
GCM-DS $\beta=0.1$     &         \bf 0.904 &    \bf 0.752 &  0.302 &           0.653 \\
GCM-DS $\beta=0.2$     &          0.899 &     0.728 &    0.332 &           0.622 \\
GCM-DS $\beta=0.5$     &          0.886 &     0.679 &    0.402 &           0.576 \\\hline
GCM-GMM $\beta=0.005$  &          0.823 &     0.449 &    0.592 &           0.402 \\
GCM-GMM $\beta=0.01$  &          0.823 &     0.487 &    0.555 &           0.362 \\
GCM-GMM $\beta=0.05$  &          0.753 &     0.586 &    0.478 &           0.179 \\
GCM-GMM $\beta=0.1$   &          0.771 &     0.550 &    0.543 &           0.214 \\
GCM-GMM $\beta=0.2$   &          0.764 &     0.513 &    0.613 &           0.199 \\
GCM-GMM $\beta=0.5$   &          0.762 &     0.488 &    0.650 &           0.194 \\\hline
\end{tabular}
} 
\label{t:supp_results_sigma0}
\end{table}
\begin{table}[h]
\centering
\caption{Different initializations of $\beta$. For SA, ARI and Scene
  Accuracy the higher the better. For VI the lower the better. Gaussian noise with $\sigma=0.1$.}
{\small
\begin{tabular}{|l|c|c|c|c|}
\hline
Model & SA $\uparrow$ & ARI $\uparrow$ & VI $\downarrow$ & Sc. Acc. $\uparrow$ \\
\hline
CCAE                &          0.754 &     0.484 &    0.689 &           0.138 \\\hline
GCM-DS $\beta=0.005$&          0.840 &     0.608 &    0.451 &           0.518 \\
GCM-DS $\beta=0.01$&          0.859 &     0.644 &    0.389 &           0.559 \\
GCM-DS $\beta=0.05$&          0.883 &     0.699 &    0.359 &       \bf    0.603 \\
GCM-DS $\beta=0.1$&      \bf    0.888 &  \bf   0.715 &  \bf  0.352 &           0.601 \\
GCM-DS $\beta=0.2$&          0.874 &     0.682 &    0.406 &           0.544 \\
GCM-DS $\beta=0.5$&          0.872 &     0.659 &    0.443 &           0.515 \\\hline
GCM-GMM $\beta=0.005$&          0.793 &     0.468 &    0.566 &           0.322 \\
GCM-GMM $\beta=0.01$&          0.800 &     0.497 &    0.547 &           0.318 \\
GCM-GMM $\beta=0.05$&          0.757 &     0.573 &    0.502 &           0.173 \\
GCM-GMM $\beta=0.1$&         0.768 &     0.521 &    0.595 &           0.204 \\
GCM-GMM $\beta=0.2$&          0.765 &     0.460 &    0.684 &           0.178 \\
GCM-GMM $\beta=0.5$&          0.753 &     0.403 &    0.780 &           0.140 \\\hline
\end{tabular}
} 
\label{t:supp_results_sigma01}
\end{table}
\begin{table}[h]
\centering
\caption{Different initializations of $\beta$. For SA, ARI and Scene
  Accuracy the higher the better. For VI the lower the better. Gaussian noise with $\sigma=0.25$.}
{\small
\begin{tabular}{|l|c|c|c|c|}
\hline
Model & SA $\uparrow$ & ARI $\uparrow$ & VI $\downarrow$ & Sc. Acc. $\uparrow$ \\
\hline
CCAE                &          0.623 &     0.248 &    0.988 &           0.033 \\\hline
GCM-DS $\beta=0.005$&          0.771 &     0.475 &    0.669 &           0.338 \\
GCM-DS $\beta=0.01$&          0.781 &     0.486 &    0.652 &           0.362 \\
GCM-DS $\beta=0.05$&    \bf      0.785 &  \bf   0.498 &    0.659 &        \bf   0.377 \\
GCM-DS $\beta=0.1$&          0.779 &     0.476 &    0.699 &           0.351 \\
GCM-DS $\beta=0.2$&          0.768 &     0.470 &    0.708 &           0.291 \\
GCM-DS $\beta=0.5$&          0.766 &     0.435 &    0.752 &           0.272 \\\hline
GCM-GMM $\beta=0.005$&          0.736 &     0.469 &    \bf 0.564 &           0.265 \\
GCM-GMM $\beta=0.01$&          0.730 &     0.467 &    0.581 &           0.221 \\
GCM-GMM $\beta=0.05$&          0.744 &     0.447 &    0.677 &           0.132 \\
GCM-GMM $\beta=0.1$&         0.735 &     0.387 &    0.801 &           0.102 \\
GCM-GMM $\beta=0.2$&          0.727 &     0.310 &    0.921 &           0.088 \\
GCM-GMM $\beta=0.5$&          0.717 &     0.242 &    1.036 &           0.064 \\\hline
\end{tabular}
} 
\label{t:supp_results_sigma025}
\end{table}

\section{Comparison across methods using the ground truth missing
indicators} \label{sec:GTmissing}.
The results above include the missing objects in
the creation of $\hat{V}$, as described in supp.\ mat.\ \ref{sec:supp_eval}.
Failing to recognize exactly which points in a
scene are missing leads to a decrease in performance for all the
metrics. However, \cite{kosiorek2019stacked} instead used the ground
truth missing indicators for the objects in order to compute the
segmentation accuracy of their method (see equation \ref{eq:SAkos}) 
If we use this strategy to all the models and metrics, we
obtain Tables~\ref{t:supp_results_gt_sigma0},
\ref{t:supp_results_gt_sigma01} and
\ref{t:supp_results_gt_sigma025}. When we compare these results with
the tables in the previous section we can see that this alternative
leads to a clear improvement in the scores of all the methods, but
with GCM-DS still performing better than CCAE or GCM-GMM.

\begin{table}[h]
\centering
\caption{Different initializations of $\beta$ using the ground truth
  missing mask. For SA, ARI and Scene Accuracy the higher the
  better. For VI the lower the better. Noise-free case.}
{\small
\begin{tabular}{|l|c|c|c|c|}
  \hline
Model & SA $\uparrow$ & ARI $\uparrow$ & VI $\downarrow$ & Sc. Acc. $\uparrow$ \\  
\hline
CCAE                &          0.950 &     0.897 &    0.157 &           0.736 \\\hline
GCM-DS $\beta=0.005$    &          0.985 &     0.967 &    0.052 &           0.915 \\
GCM-DS $\beta=0.01$   &      \bf    0.989 &  \bf   0.973 &  \bf  0.042 &       \bf    0.941 \\
GCM-DS $\beta=0.05$   &          0.971 &     0.938 &    0.095 &           0.886 \\
GCM-DS $\beta=0.1$  &          0.966 &     0.929 &    0.108 &           0.860 \\
GCM-DS $\beta=0.2$  &          0.966 &     0.930 &    0.104 &           0.854 \\
GCM-DS $\beta=0.5$  &          0.958 &     0.912 &    0.132 &           0.825 \\\hline
GCM-GMM $\beta=0.005$   &          0.886 &     0.805 &    0.209 &           0.646 \\
GCM-GMM $\beta=0.01$  &          0.905 &     0.830 &    0.191 &           0.683 \\
GCM-GMM $\beta=0.05$  &          0.942 &     0.884 &    0.160 &           0.718 \\
GCM-GMM $\beta=0.1$ &          0.937 &     0.879 &    0.179 &           0.657 \\
GCM-GMM $\beta=0.2$ &          0.927 &     0.861 &    0.213 &           0.603 \\
GCM-GMM $\beta=0.5$ &          0.925 &     0.858 &    0.216 &           0.585 \\\hline
\end{tabular}
} 
\label{t:supp_results_gt_sigma0}
\end{table}
\begin{table}[h]
\centering
\caption{Different initializations of $\beta$ using the ground truth
  missing mask. For SA, ARI and Scene Accuracy the higher the
  better. For VI the lower the better. Gaussian noise with $\sigma=0.1$}
{\small
\begin{tabular}{|l|c|c|c|c|}
  \hline
Model & SA $\uparrow$ & ARI $\uparrow$ & VI $\downarrow$ & Sc. Acc. $\uparrow$ \\  
\hline
CCAE                &          0.867 &     0.729 &    0.403 &           0.386 \\\hline
GCM-DS $\beta=0.005$    &          0.957 &     0.917 &    0.135 &           0.794 \\
GCM-DS $\beta=0.01$   &     \bf     0.972 &   \bf  0.940 & \bf   0.094 &     \bf      0.857 \\
GCM-DS $\beta=0.05$   &          0.957 &     0.908 &    0.142 &           0.833 \\
GCM-DS $\beta=0.1$  &          0.956 &     0.906 &    0.144 &           0.831 \\
GCM-DS $\beta=0.2$  &          0.946 &     0.887 &    0.172 &           0.794 \\
GCM-DS $\beta=0.5$  &          0.942 &     0.875 &    0.194 &           0.759 \\\hline
GCM-GMM $\beta=0.005$   &          0.856 &     0.752 &    0.272 &           0.544 \\
GCM-GMM $\beta=0.01$  &          0.875 &     0.768 &    0.272 &           0.561 \\
GCM-GMM $\beta=0.05$  &          0.922 &     0.837 &    0.232 &           0.621 \\
GCM-GMM $\beta=0.1$ &          0.912 &     0.824 &    0.268 &           0.548 \\
GCM-GMM $\beta=0.2$ &          0.905 &     0.822 &    0.278 &           0.474 \\
GCM-GMM $\beta=0.5$ &          0.881 &     0.792 &    0.334 &           0.379 \\\hline
\end{tabular}
} 
\label{t:supp_results_gt_sigma01}
\end{table}
\begin{table}[h]
\centering
\caption{Different initializations of $\beta$ using the ground truth
  missing mask. For SA, ARI and Scene Accuracy the higher the
  better. For VI the lower the better. Gaussian noise with $\sigma=0.25$}
{\small
\begin{tabular}{|l|c|c|c|c|}
  \hline
Model & SA $\uparrow$ & ARI $\uparrow$ & VI $\downarrow$ & Sc. Acc. $\uparrow$ \\  
\hline
CCAE                &          0.774 &     0.621 &    0.540 &           0.033 \\\hline
GCM-DS $\beta=0.005$    &          0.897 &     0.811 &    0.302 &           0.556 \\
GCM-DS $\beta=0.01$   &      \bf    0.908 & \bf    0.822 &  \bf  0.281 &           0.611 \\
GCM-DS $\beta=0.05$   &          0.900 &     0.805 &    0.302 &       \bf    0.623 \\
GCM-DS $\beta=0.1$  &          0.895 &     0.796 &    0.317 &           0.609 \\
GCM-DS $\beta=0.2$  &          0.891 &     0.784 &    0.333 &           0.578 \\
GCM-DS $\beta=0.5$  &          0.881 &     0.770 &    0.359 &           0.528 \\\hline
GCM-GMM $\beta=0.005$   &          0.852 &     0.743 &    0.287 &           0.545 \\
GCM-GMM $\beta=0.01$  &          0.865 &     0.756 &    0.282 &           0.558 \\
GCM-GMM $\beta=0.05$  &          0.887 &     0.773 &    0.337 &           0.468 \\
GCM-GMM $\beta=0.1$ &          0.869 &     0.748 &    0.392 &           0.355 \\
GCM-GMM $\beta=0.2$ &          0.855 &     0.734 &    0.419 &           0.278 \\
GCM-GMM $\beta=0.5$ &          0.834 &     0.704 &    0.470 &           0.196 \\\hline
\end{tabular}
} 
\label{t:supp_results_gt_sigma025}
\end{table}

\section{Examples of CCAE and GCM-DS inference for the constellations data}
\label{sec:ccae_gcmds_inferences}

\begin{figure*}[t]
  \centering
  \caption{Reconstruction examples from CCAE and GCM-DS for noise-free
    data. The upper figures show the ground truth of the test images. The middle
    figures show the reconstruction and the capsule assignments (by
    colours) of CCAE. The lower figures show the
    reconstruction and the parts assignment of GCM-DS. Datapoints
      shown with a given colour are predicted
    to belong to the reconstructed object with the corresponding colour.}
  \includegraphics[width=0.7\linewidth]{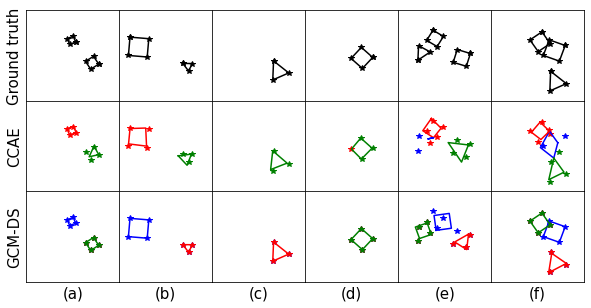}
  \label{fig:examples}
\end{figure*}

Figure~\ref{fig:examples} shows some reconstruction examples from CCAE
and GCM-DS for the noise-free scenario.  We show additional examples
for the noisy cases in the supp.\ mat.\ \ref{sec:noisy_examples}.  In
columns (a) and (b) we can see that CCAE recovers the correct parts
assignments but the object reconstruction is amiss.  In (a) one of the
squares is reconstructed as a triangle, while in (b) the assignment
between the reconstruction and the ground truth is not exact.  For
GCM-DS, if the parts are assigned to the ground truth properly, {and
  there is no noise}, then the reconstruction of the object is
perfect.  In column (c) all methods work well.  In column (d), CCAE
fits the square correctly (green), but adds an additional red point.
In this case GCM-DS actually overlays two squares on top of each
other.  Both methods fail badly on column (e).  Note that CCAE is not
guaranteed to reconstruct an existing object correctly (square or
triangle in this case).  In column (f) we can see that CCAE fits an
irregular quadrilateral (blue) to the assigned points, while GCM-DS
obtains the correct fit.

\section{Noisy cases examples}\label{sec:noisy_examples}

In Figures~\ref{fig:examples_sigma01} and
  \ref{fig:examples_sigma025} we show several examples of objects
  generated from noisy templates {with corruption levels of
  $\sigma=0.1$ and $\sigma=0.25$ resp.}
GCM-DS and RANSAC does not allow for deformable objects to try to fit
the points exactly, contrary to CCAE.
Both methods try to find the closest reconstruction of the noisy points in the image by selecting the geometrical shapes that are a best fit to those points.
Nonetheless, both methods can determine that which parts belong
together to form a given object, even when the matching is not
perfect.

\begin{figure*}[t]
  \centering
  \caption{Reconstruction examples from CCAE, GCM-DS and RANSAC with Gaussian noise $\sigma=0.1$.}
  \includegraphics[width=0.7\linewidth]{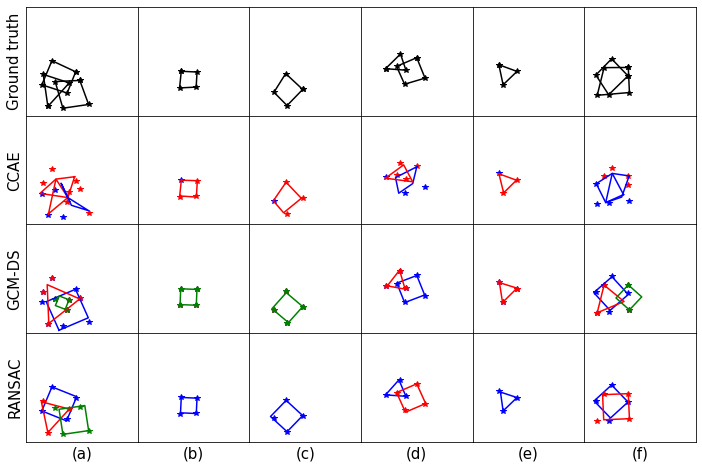}
  \label{fig:examples_sigma01}
\end{figure*}

\begin{figure*}[t]
  \centering
  \caption{Reconstruction examples from CCAE, GCM-DS and RANSAC with Gaussian noise $\sigma=0.25$.}
  \includegraphics[width=0.7\linewidth]{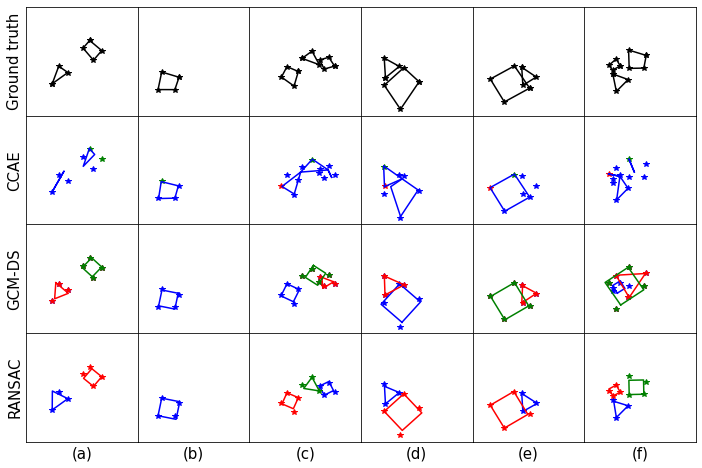}
  \label{fig:examples_sigma025}
\end{figure*}

\begin{figure*}[t]
    \centering
    \caption{Average part-based reconstruction for different angles of rotation of the input scenes. Row (a) corresponds to reconstructions with the learned part-sets 2 and row (b) reconstructions with the learned part-sets 3 of Fig. \ref{fig:supp_templates}.}
    \begingroup
    \setlength{\tabcolsep}{1pt} 
    \renewcommand{\arraystretch}{1} 

    \begin{tabular}{c cccccccc}
    
    (a) &
    \includegraphics[align=c,scale=0.2]{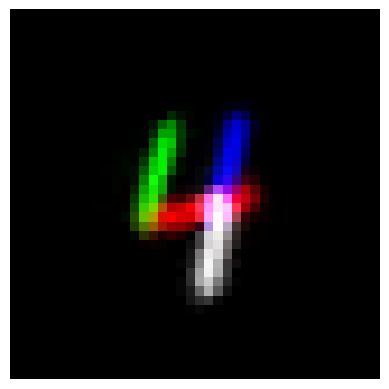} &
    \includegraphics[align=c,scale=0.2]{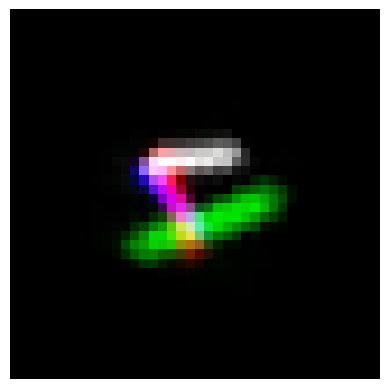} & 
    \includegraphics[align=c,scale=0.2]{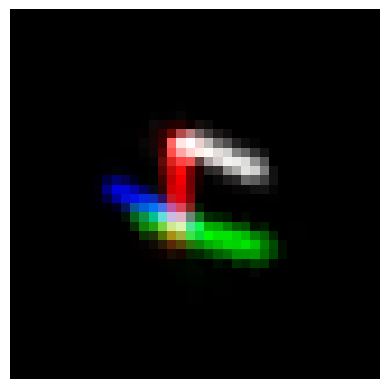} &
    \includegraphics[align=c,scale=0.2]{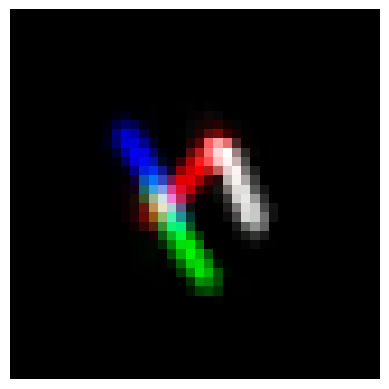} &
    \includegraphics[align=c,scale=0.2]{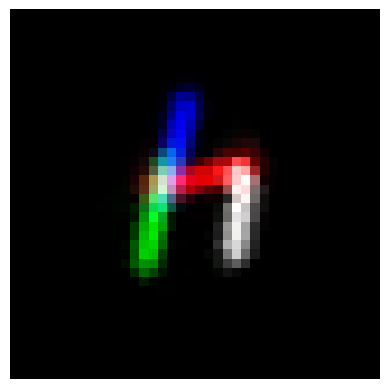} &
    \includegraphics[align=c,scale=0.2]{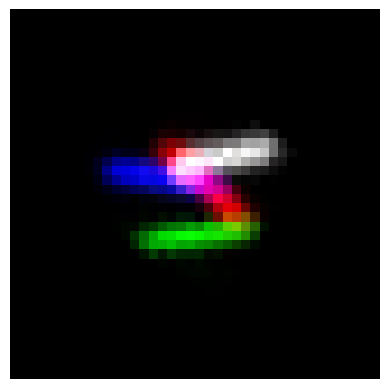} &
    \includegraphics[align=c,scale=0.2]{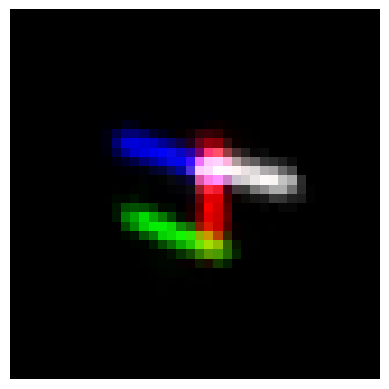} &
    \includegraphics[align=c,scale=0.2]{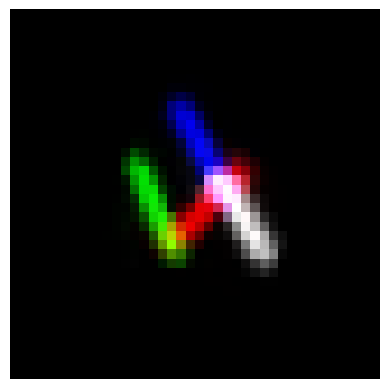} \\
    
    (b) &
    \includegraphics[align=c,scale=0.2]{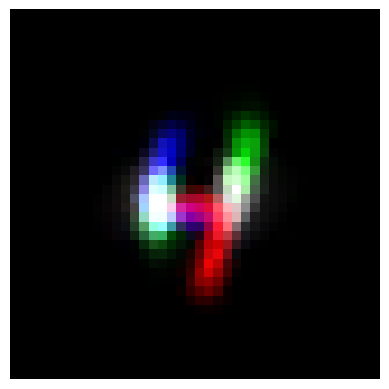} &
    \includegraphics[align=c,scale=0.2]{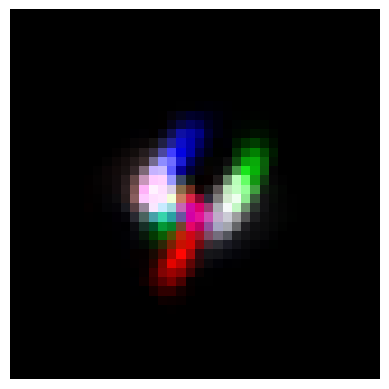} & 
    \includegraphics[align=c,scale=0.2]{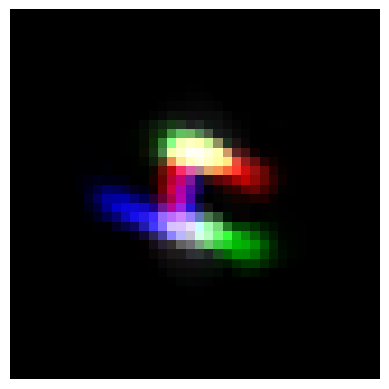} &
    \includegraphics[align=c,scale=0.2]{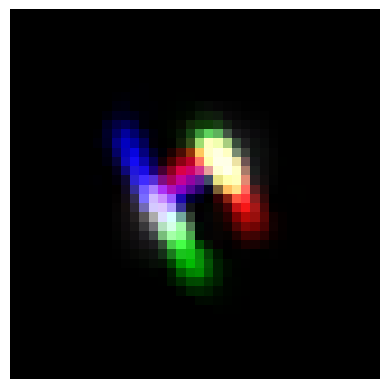} &
    \includegraphics[align=c,scale=0.2]{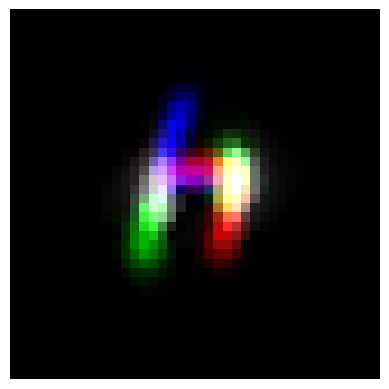} &
    \includegraphics[align=c,scale=0.2]{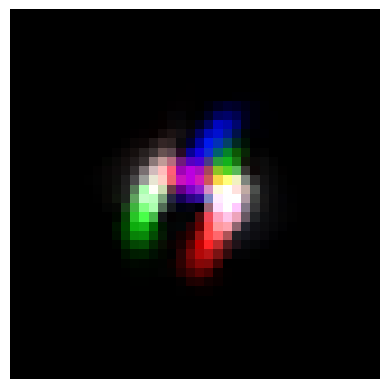} &
    \includegraphics[align=c,scale=0.2]{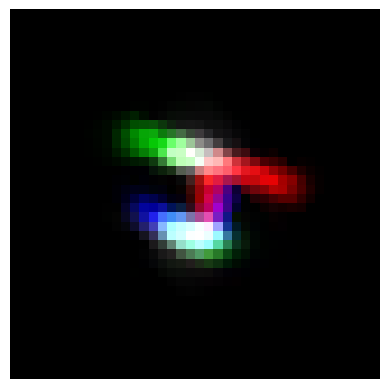} &
    \includegraphics[align=c,scale=0.2]{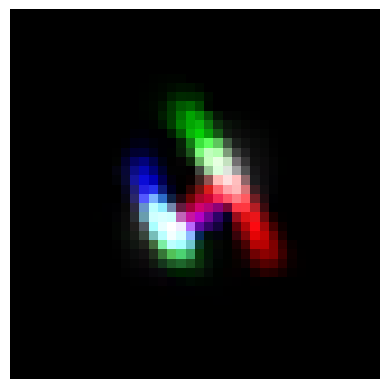} \\
    
    & $0^\circ$ &$45^\circ$ &$90^\circ$ &$135^\circ$ &$180^\circ$ &$225^\circ$ &$270^\circ$ &$315^\circ$ \\

    \end{tabular}
    \endgroup
    \label{fig:supp_PCAE_equiv}
\end{figure*}

\begin{figure*}[t]
  \centering
  \caption{Reconstruction examples with our Variational Inference (VI) algorithm and the RANSAC-type algorithm: (a) scene with 2 faces, (b) scene with 3 faces, (c) scene with 4 faces (d) scene with 5 faces and (e) 3 faces with partially occluded faces. All faces have been randomly selected and transformed.}
\begingroup
\setlength{\tabcolsep}{1pt} 
\renewcommand{\arraystretch}{1} 
\begin{tabular}{c ccc}
    Ground Truth & VI & RANSAC \\
    (a) \includegraphics[align=c,scale=0.3]{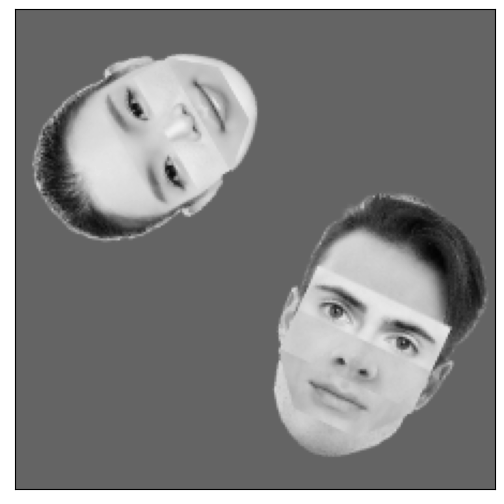} &
    \includegraphics[align=c,scale=0.3]{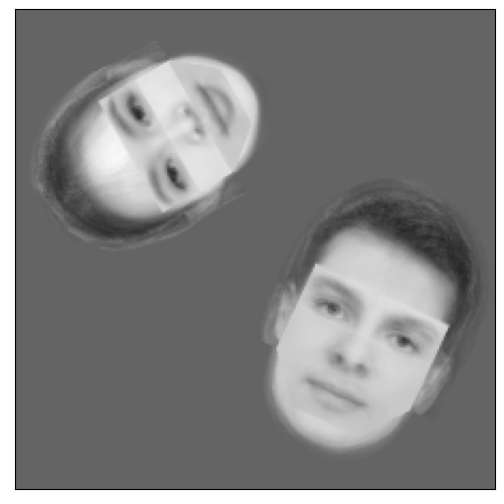} &
    \includegraphics[align=c,scale=0.3]{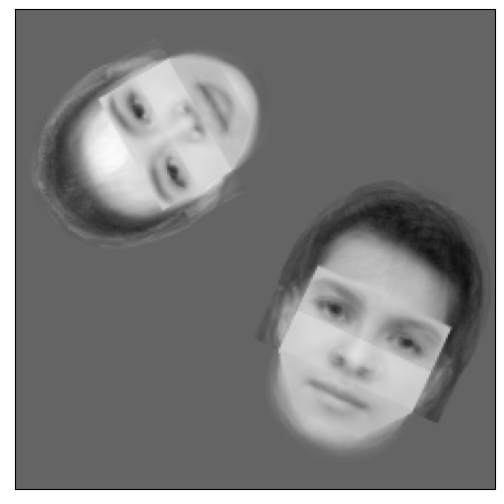} \\

    (b) \includegraphics[align=c,scale=0.3]{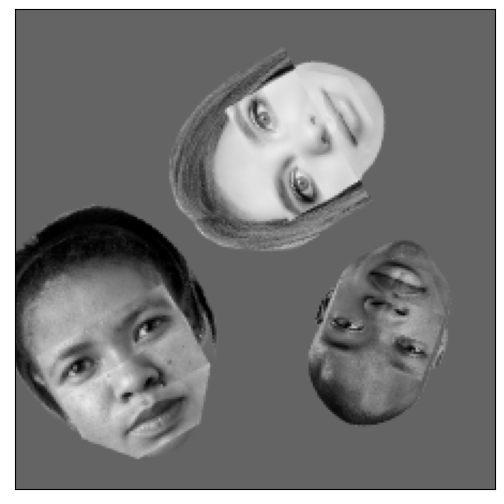} &
    \includegraphics[align=c,scale=0.3]{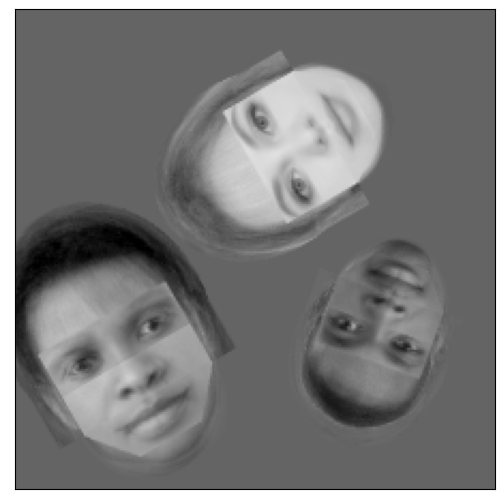} &
    \includegraphics[align=c,scale=0.3]{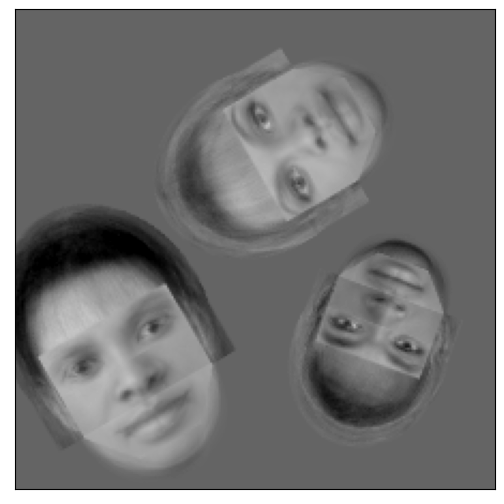} \\

    (c) \includegraphics[align=c,scale=0.3]{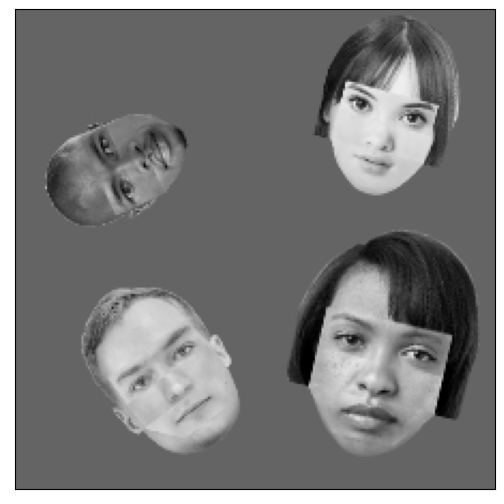} &
    \includegraphics[align=c,scale=0.3]{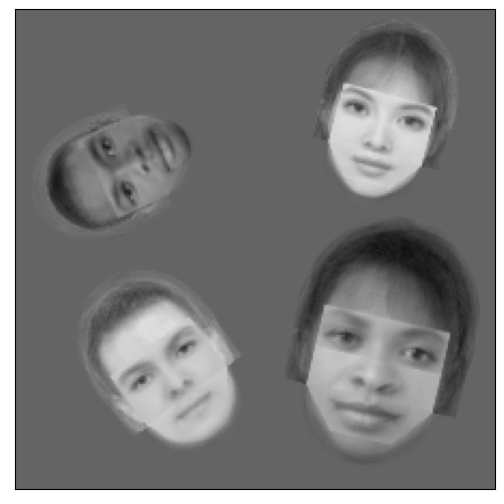} &
    \includegraphics[align=c,scale=0.3]{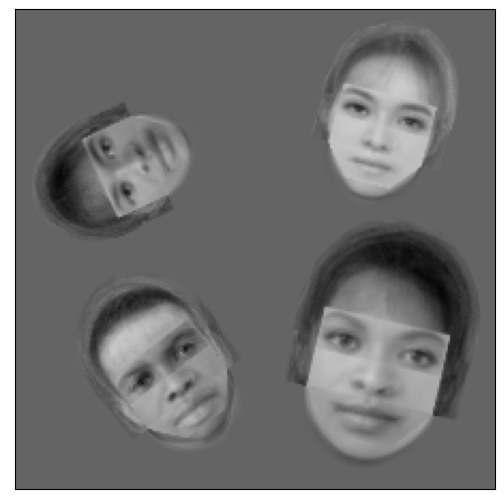} \\
 
    (d) \includegraphics[align=c,scale=0.3]{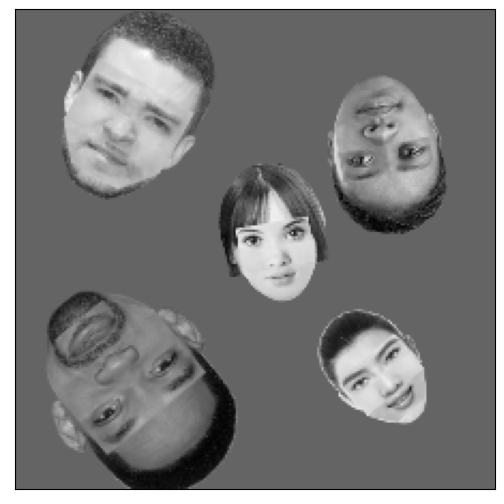} &
    \includegraphics[align=c,scale=0.3]{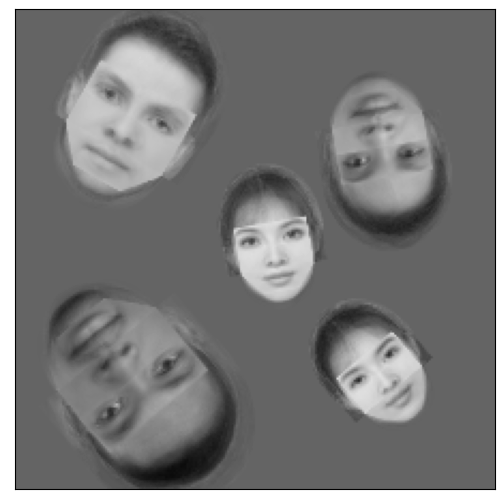} &
    \includegraphics[align=c,scale=0.3]{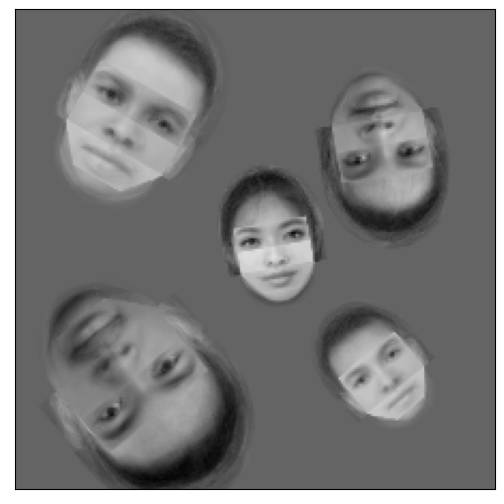} \\
    
    (e) \includegraphics[align=c,scale=0.3]{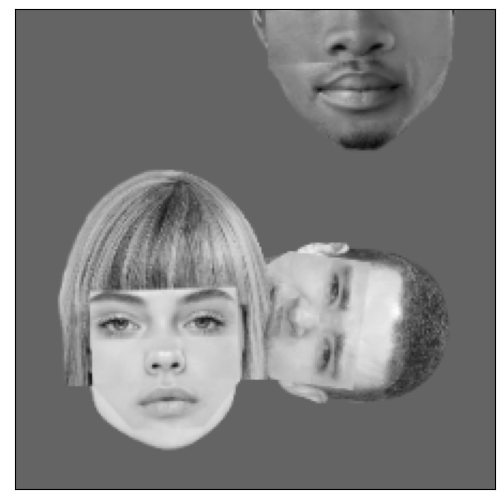} &
    \includegraphics[align=c,scale=0.3]{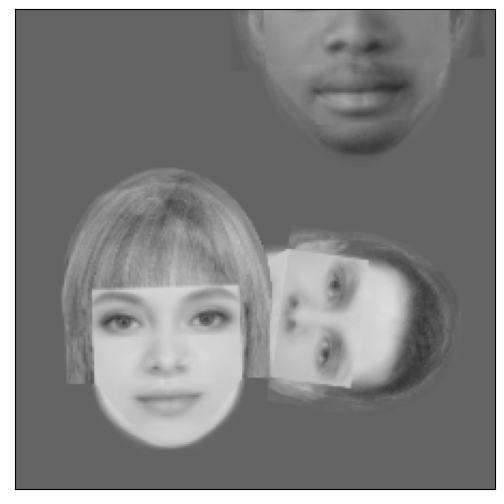} &
    \includegraphics[align=c,scale=0.3]{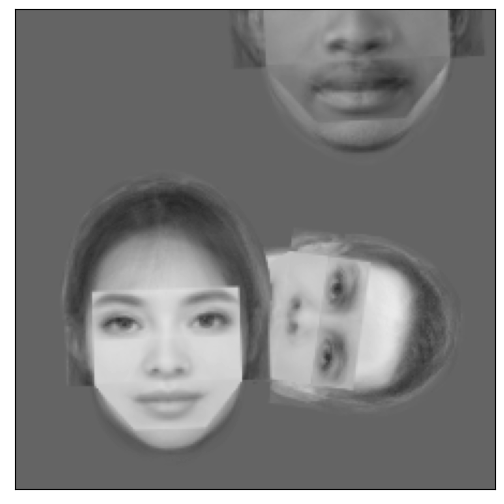} \\
      \end{tabular}
\endgroup
\label{fig:face_inference_examples}
\end{figure*}

\end{document}